\documentclass[final,12pt]{clear2025}

\usepackage{bbm}
\usepackage{tcolorbox}
\usepackage{amsmath}
\usepackage{listings}
\usepackage{cancel}
\usepackage{tikz}
\usetikzlibrary{positioning}
\usepackage{natbib}
\usepackage{booktabs}
\usetikzlibrary{graphs}
\usepackage{makecell}
\usepackage{amssymb}
\usepackage{float}
\usepackage{bm}
\usepackage{mathtools} 
\usepackage{changepage}

\usepackage{rotating}
\usepackage{multirow}
\usepackage{booktabs}

\newcommand{\allvars}{\mathbf{V}}

\newcommand{\inputvars}{\vars^{\mathbf{In}}}

\newcommand{\outputvars}{\vars^{\mathbf{Out}}}

\newcommand{\tset}{\mathbf{v}}

\newcommand{\values}[1]{\mathsf{Val}_{#1}}
\newcommand{\signature}{\Sigma}
\newcommand{\gpt}{GPT 2-small}

\newcommand{\onevar}{X}
\newcommand{\onevarr}{Y}

\newcommand{\onevall}{y}

\newcommand{\vars}{\mathbf{X}}

\newcommand{\varrrs}{\mathbf{Z}}

\newcommand{\pset}{\mathbf{x}}

\newcommand{\psettt}{\mathbf{z}}

\newcommand{\simfunc}{\mathsf{Sim}}
\newcommand{\statistic}{\mathbb{S}}

\newcommand{\base}{\mathbf{b}}
\newcommand{\source}{\mathbf{s}}

\newcommand{\dintinv}{\mathsf{DistIntInv}}
\newcommand{\intinv}{\mathsf{IntInv}}

\newcommand{\project}[2]{\mathsf{Proj}_{#2}(#1)}

\newcommand{\inverseproject}[2]{\mathsf{Proj}^{-1}_{#2}(#1)}

\newcommand{\mechanisms}[2]{\{\mathcal{F}_{#1}\}_{#1 \in #2}}

\newcommand{\mechanism}[1]{\mathcal{F}_{#1}}

\newcommand{\model}{\mathcal{M}}

\newcommand{\intpset}{\mathbf{i}}

\newcommand{\intvars}{\mathbf{I}}

\newcommand{\inmap}{\delta}
\newcommand{\setmap}{\tau}
\newcommand{\intmap}{\omega}

\newcommand{\cellpart}{\Pi}
\newcommand{\cellfunc}{\pi}

\newcommand{\lowmodel}{\mathcal{L}}
\newcommand{\highmodel}{\mathcal{H}}

\newcommand{\distribution}{\mathbb{P}}

\newcommand{\bx}{\mathbf{b}}
\newcommand{\sx}{\mathbf{s}}

\title[Combining Causal Models for More Accurate Abstractions]{Combining Causal Models for\\ More Accurate Abstractions of Neural Networks}
\usepackage{times}

\clearauthor{%
 \Name{Theodora-Mara Pîslar} \Email{pislarmara589@gmail.com}\\
 \addr University of Amsterdam
 \AND
 \Name{Sara Magliacane} \Email{sara.magliacane@gmail.com}\\
 \addr University of Amsterdam
 \AND
 \Name{Atticus Geiger} \Email{atticusg@gmail.com}\\
 \addr Pr(Ai)²R Group%
}

\begin{document}

\maketitle

\begin{abstract}%
\textit{Mechanistic interpretability} aims to reverse engineer neural networks by uncovering which high-level algorithms they implement. Causal abstraction provides a precise notion of when a network implements an algorithm, i.e., a causal model of the network contains low-level features that realize the high-level variables in a causal model of the algorithm \citep{geiger2024causalabstractiontheoreticalfoundation}. A typical problem in practical settings is that the algorithm is not an entirely faithful abstraction of the network, meaning it only \emph{partially} captures true reasoning process of a model. We propose a solution where we \textit{combine} different simple high-level models to produce a more faithful representation of the network. Through learning this combination, we can model neural networks as being in different computational states depending on the input provided, which we show is more accurate to \gpt\ fine-tuned on two toy tasks.
We observe a trade-off between the \textit{strength} of an interpretability hypothesis, which we define in terms of the number of inputs explained by the high-level models, and its \textit{faithfulness}, which we define as the interchange intervention accuracy. Our method allows us to modulate between the two, providing the most accurate combination of models that describe the behavior of a neural network given a faithfulness level. The code is available in \href{https://github.com/maraPislar/combining-causal-models-for-accurate-NN-abstractions}{github}.
\end{abstract}

\section{Introduction}
Great strides have been made in gaining a mechanistic understanding of black box deep learning models using tools from causal mediation \citep{vig2020causal, finlayson-etal-2021-causal, mueller2024questrightmediatorhistory} and causal abstraction \citep{geiger-etal-2020-neural,geiger2021causal, geiger2024causalabstractiontheoreticalfoundation, huang2024ravel}. However, an important area of innovation is quantifying and reasoning about the \textit{faithfulness} of a causal analysis of a deep learning model, i.e., the degree to which the abstract causal model accurately represents the true reasoning process of a model \citep{Lipton18, Jacovi2020, Lyu}.

In this paper, we explore an approach in which a causal model that is a partially faithful description of a neural network is modified to become more faithful by dynamically changing the causal structure based on what input is provided. We represent algorithms as causal models and define a \textit{combine} operation that creates a causal model with a mechanism that activates different variables based on the input provided, with some inputs activating no intermediate variables. These combined models are more expressive than the original causal models, allowing for more fine-grained hypotheses about the computational process. 

However, there is a trade-off between faithfulness and the strength of an interpretability hypothesis. We measure strength by the number of examples that a combination of causal models explains at a given level of faithfulness, i.e., examples assigned to a causal model with internal structure. A causal model with no internal structure has no content and is a trivially faithful hypothesis about network structure; the more interesting structure a causal model has, the more chances there are for inaccuracies. In two experiments with \gpt\ on arithmetic and boolean logic tasks, we show that combined causal models are able to provide stronger hypotheses at every level of faithfulness.

\section{Background}\label{sec:back}
This section covers previous research on causal abstraction and interpretability.

\paragraph{Mechanistic Interpretability.}
Mechanistic interpretability has converged on analyzing distributed representations, with significant attention to linear subspaces following the linear representation hypothesis \citep{mikolov-etal-2013-linguistic, elhage2022superposition, Nanda2023, Park:2023, Jiang:2024}. Methods like Distributed Alignment Search \citep{Geiger-etal:2023:DAS, Wu:Geiger:2023:BDAS} and sparse autoencoders \citep{bricken2023monosemanticity, Cunningham:2023} have proven effective in identifying interpretable features, while intervention techniques provide tools for validating interpretability hypotheses, e.g., interchange interventions \citep{geiger-etal-2020-neural, vig2020causal}, path patching \citep{wang2023interpretability, goldowskydill2023localizing}, activation patching \citep{automated_discovery, zhang2024towards, patchscope}, and causal scrubbing \citep{chan2022causal}. Circuit analysis has revealed specific computational mechanisms across both visual \citep{curve_circuits, zoomin} and linguistic domains \citep{wang2023interpretability, olsson2022context, Lieberum:2023}. Recent theoretical frameworks \citep{geiger2024causalabstractiontheoreticalfoundation, mueller2024questrightmediatorhistory} and evaluation methods \citep{huang2024ravel} are establishing more rigorous standards for mechanistic explanations.

\paragraph{Faithfulness in Interpretability.}
Faithfulness is a critical concept in interpretability research, referring to the degree to which an explanation accurately represents the true reasoning process of a model \citep{Jacovi2020, Lyu}. A faithful interpretation should not only match model behavior, but also reflect the internal decision-making process \citep{notnotAttention}. This is particularly important in high-stakes domains where understanding the model's reasoning is crucial for trust and accountability. However, measuring faithfulness is challenging, as it often requires comparing explanations against ground truth reasoning processes, which are typically unknown for complex models. Despite advances, achieving truly faithful interpretations remains an open challenge in the field of explainable AI \citep{Lipton18}.

\paragraph{Causal Models.}
We adopt the following notation and concepts from \cite{Bongers} and \cite{geiger2024causalabstractiontheoreticalfoundation}.
We define a deterministic causal model $\model = (\signature,\mechanisms{\onevar}{\allvars})$, where $\signature = (\allvars ,\values{})$ is a signature consisting of a set of variables $\allvars$ and their corresponding value ranges $\values{}$. The mechanisms $\mechanisms{\onevar}{\allvars}$ assign a value to each variable $\onevar$ as a function of all variables, including itself.
We limit ourselves to acyclic models with input variables $\inputvars$ that depend on no other variables and output variables $\outputvars$ on which no other variables depend. The remaining variables are considered intermediate variables. 
The solution $\model(\pset)$ of a causal model given an input setting $\pset$ is the unique total setting that satisfies the mechanisms for each $\onevar \in \allvars$, i.e., $\project{\tset}{\onevar}$, the projection of $\tset \in \values{\allvars}$ onto $\onevar$, is equal to the mechanism output $\mechanism{\onevar}(\tset)$.
While our definition does not explicitly reference a graphical structure, the mechanisms induce a causal ordering $\prec$ among variables, where $\onevarr \prec \onevar$ if there exists a setting $\psettt$ of variables $\varrrs = \allvars\setminus\{\onevarr\}$ and two settings $\onevall,\onevall'$ of $\onevarr$ such that $\mechanism{\onevar}(\psettt,\onevall) \neq \mechanism{\onevar}(\psettt,\onevall')$.

\paragraph{Interventions.}
Interventions are operations on the mechanisms of a causal model. A hard intervention $\intpset \in  \values{\intvars}$ for $\intvars \subseteq \allvars$ replaces the mechanisms $\mechanism{\onevar}$ for each $\onevar \in \intvars$ with constant functions $\tset \mapsto \project{\intpset}{\onevar}$. The causal model resulting from an intervention $\gamma$ is denoted as $\model_{\gamma}$.

Interventionals generalize interventions to arbitrary mechanism transformation, i.e., a functional that outputs a new mechanism conditional on the mechanisms of the original model \citep{geiger2024causalabstractiontheoreticalfoundation}. We will not focus on interventionals, though they are necessary for a rigorous treatment of abstraction where high-level variables are aligned to linear subspaces of low-level variables.

\paragraph{Exact Transformation.}
Exact transformation is a fundamental concept \citep{Rubinstein2017} that formalizes under which conditions are two causal models $\model, \model^*$ are compatible representations of the same causal phenomena. Exact transformation holds with respect to three maps $\inmap, \setmap, \intmap$ defined on inputs, total settings that assign each variable a value, and interventions, respectively, \footnote{Previous definitions did not use an input map, we include one here for simplicity's sake.} if for each input $\base$ and intervention $\intpset$, we have:
\[\setmap\Big(\model_{\intpset}(\base)\Big)  = \model^*_{\intmap(\intpset)}\Big(\inmap(\base)\Big).\]
This equation holds exactly when every low-level intervention $\intpset$ applied to the low-level model $\model$ is mapped by $\setmap$ to the same high-level setting that results from the low-level intervention $\intpset$ being mapped to a high-level intervention via $\intmap(\intpset)$  and then applied to the high-level model $\model^*$. \cite{geiger2024causalabstractiontheoreticalfoundation} generalize this notion to \textit{intervention algebras}, which are sets of interventionals that fix quantities distributed across variables, e.g., a linear subspace of a vector of variables.

\paragraph{Constructive Causal Abstraction.}
Exact transformation as a general notion is unconstrained; there is no guaranteed relationship between high-level variables and low-level variables. Constructive causal abstraction \citep{BackersHalpern2019,beckers20a, Massidda2023, RischelWeichwald} is a special case of exact transformation between a low-level model $\lowmodel$ and a high-level model $\highmodel$.
An alignment $\langle \Pi, \pi \rangle$ assigns each high-level variable $\onevar \in \allvars_{\highmodel}$ a partition cell $\cellpart_{\onevar} \subseteq \allvars_{\lowmodel}$ of low-level variables and a function $\cellfunc_{\onevar}$ for determining the value of $\onevar$ from a setting of $\cellpart_{\onevar}$. From an alignment, we can construct a map $\setmap^{\cellfunc}$ from low-level total settings $\tset_{\lowmodel}$ to high-level total settings:
\[\setmap^{\cellfunc}(\tset_{\lowmodel}) = \bigcup_{\onevar_{\highmodel} \in \allvars_{\highmodel}} \cellfunc_{\onevar_{\highmodel}} \big( \project{\tset_{\lowmodel}}{\cellpart_{\onevar_{\highmodel}}} \big),\]
where $\project{\tset_{\lowmodel}}{\cellpart_{\onevar_{\highmodel}}}$ represents the projection of the low-level total settings to the values for the low-level variables $\cellfunc_{\onevar_{\highmodel}}$. 
We can also construct a map $\intmap^{\cellfunc}$ from low-level hard interventions to high-level hard interventions
where we define $\intmap^{\cellfunc}(\pset_{\lowmodel}) = \pset_{\highmodel}$ iff:
\[\setmap^{\cellfunc}\big(\inverseproject{\pset_{\lowmodel}}{ \allvars_{\lowmodel}}\big) = \inverseproject{\pset_{\highmodel}}{ \allvars_{\highmodel}}.\]
A model $\highmodel$ is a constructive abstraction of a model $\lowmodel$ iff $\highmodel$ is an exact transformation of $\lowmodel$ under $(\inmap, \setmap^{\cellfunc}, \intmap^{\cellfunc})$. Constructive abstraction is not defined on interventionals, and an analysis where high-level variables are aligned to linear subspaces requires an exact transformation of the network to change the basis and the algorithm is a constructive causal abstraction of this transformed model. 

\paragraph{Applied Causal Abstraction.} Causal abstraction has been used to analyze weather patterns \citep{chalupka2016}, human brains \citep{Dubois2020PersonalityBT,Dubois2020CausalMO}, physical systems \citep{Kekic2023}, batteries \citep{Zennaro23}, epidemics \citep{Dyer2023}, and deep learning models \citep{Chalupka:2015, geiger2021causal, hu22b, Geiger-etal:2023:DAS, Wu:Geiger:2023:BDAS}. 

\section{Mechanistic Interpretability via Causal Abstraction}
The core of mechanistic interpretability is reverse engineering the algorithms implemented by a neural network to solve a task. We operationalize a mechanistic interpretability hypothesis about internal structure by representing neural networks and algorithms both as causal models and then understanding implementation as a type of exact transformation. However, to uncover the structure of a neural network, we need to \textit{featurizes} hidden representations. We use the high-level causal model as a source of supervision to learn a bijective exact transformation that creates groups of orthogonal linear features corresponding to variables in a high-level neural network.

\paragraph{Interchange Interventions.}
The theory of causal abstraction does not specify \textit{which} interventions should be evaluated. \cite{geiger2021causal} argue for using interventions that fix variables to the values they would have if some counterfactual input were provided, as such low-level counterfactual states have meanings determined fully by the input and the high-level model. Given a model $\model$ with counterfactual input $\source$ and target variables $\vars$, a single source interchange intervention is $\intinv(\model, \source, \vars) = \project{\model(\source)}{\vars}$, the value that $\vars$ takes on when $\source$ is input.

Distributed interchange interventions (DII) generalize standard interchange interventions to target quantities that are distributed across several causal variables. This is crucial for analyzing complex neural networks where individual neurons often participate in multiple conceptual roles. A bijective function is applied before the interchange intervention and the inverse is applied after. A distributed interchange intervention is an interventional, i.e., rather than fixing variables to constant values, the variable mechanisms are replaced with a new set of mechanisms.

For our purposes, the bijective function is a rotation that allows interventions on the dimensions of a new basis. The hyperparameter $k$ determines the number of dimensions we will intervene on.  Given a model $\model$, counterfactual input $\sx$, hidden representation $\mathbf{H} \in \mathbb{R}^n$, and a matrix with $k$ orthogonal columns $\mathbf{R} \in \mathbb{R}^{n \times k}$, the distributed interchange intervention is the intervention $\dintinv(\model, \sx, \mathbf{H}, \mathbf{R})$ that fixes the linear subspace spanned by the columns of $\mathbf{R}$ to the value they take for counterfactual input $\sx$. The mechanism of $\model_{\dintinv(\model, \sx, \mathbf{H}, \mathbf{R})}$ for the variables $\mathbf{H}$ is 
\begin{equation}
\tset \mapsto \mechanism{\mathbf{H}}(\tset)+ \mathbf{R}^{T}\Bigg(\mathbf{R}\Big(\mechanism{\mathbf{H}}\big(\model(\sx)\big)\Big) - \mathbf{R}\big(\mechanism{\mathbf{H}}(\tset)\big)\Bigg)
\end{equation}
This new mechanism maps the total setting $\tset$ to the original mechanism, except the subspace spanned by $\mathbf{R}$ is erased by subtracting $\mathbf{R}\big(\mechanism{\mathbf{H}}(\tset)\big)$ and then fixed to be $\mathbf{R}\Big(\mechanism{\mathbf{H}}\big(\model(\sx)\big)\Big)$,  the value that the subspace would take on when the counterfactual input $\mathbf{s}$ is run through the model $\mathcal{M}$.

\paragraph{Distributed Alignment Search.}

Distributed Alignment Search (\citealt{Geiger-etal:2023:DAS}; DAS) is a method for aligning linear subspaces of a hidden representation with high-level variables. DAS optimizes an orthogonal matrix $\mathbf{R}$ using distributed interchange interventions. Given a fixed alignment, define a loss function for each high-level variable $\onevar$ and low-level variables $\cellpart_{\onevar}$ with rotation matrix $\mathbf{R}_{\onevar}$. Given a base input $\bx$, counterfactual input $\sx$, and high-level model $\highmodel$, the loss is
\begin{equation}
\mathcal{L}_{\mathsf{DAS}}=\sum \mathsf{CE}\big(\lowmodel_{\dintinv(\lowmodel, \sx, \cellpart_{\onevar}, \mathbf{R}_X)}(\bx), \highmodel_{\intinv(\highmodel, \source, \onevar)}(\base)\big)
\end{equation}
where CE is the cross-entropy loss.

\paragraph{Approximate Transformation and Interchange Intervention Accuracy.}

While exact transformation provides a strict criterion for abstraction, in practice we quantify the degree to which a high-level model approximates a low-level model. \cite{geiger2024causalabstractiontheoreticalfoundation} generalize exact transformation to approximate transformation by introducing a similarity function $\simfunc$ between total settings, a probability distribution $\distribution$ over interventions, and a statistic $\statistic$ to aggregate similarity scores. 
Interchange Intervention Accuracy (IIA) is a specific instance of approximate abstraction that measures the proportion of interchange interventions where the low-level and high-level models produce the same output. IIA for a single high-level variable $\onevar$ over counterfactual dataset $\mathcal{D}$ is defined as:
\begin{equation}
\mathsf{IIA}(\highmodel, \lowmodel, \inmap, \setmap, \intmap, \mathcal{D}) =\frac{1}{|\mathcal{D}|}\sum_{\base, \source \in \mathcal{D}} \mathbbm{1}\big[\lowmodel_{\dintinv(\lowmodel, \sx, \cellpart_{\onevar}, \mathbf{R}_X)}(\bx) = \highmodel_{\intinv(\highmodel, \source, \onevar)}(\base)\big]
\end{equation}
Interchange intervention accuracy will serve as our faithfulness metric quantifying the degree to which a high-level causal model $\mathcal{H}$ is a causal abstraction of a low-level neural network $\mathcal{L}$. Appendix \ref{sec:counterfactual_data} shows a step-by-step example of how a counterfactual dataset is constructed.

\section{Preliminary Interpretability Analysis on a Simple Arithmetic Task}\label{sec:arith}
We will motivate the approach in this work by demonstrating the issues of partially faithful abstractions. Our running example is the task of summing three numbers $X$, $Y$, and $Z$ that vary from 1 to 10. We fine-tune \gpt\ to perfectly solve the task, but the fundamental problem of interpretability is that deep learning models are black boxes and we do not know \textit{how} the model solves the task. However, this is a simple task and we can easily enumerate some algorithms that solve it.\footnote{Obviously, there are an infinite number of algorithms that solve any input-output task, as useless computational structure can be added ad infinitum. Nonetheless, a simple input-output task makes this space easier to think about and search through.}

\subsection{Interpretability Hypotheses} We consider five different states that \gpt\ might be in when adding three numbers $X, Y, Z$. One initial state where $X$, $Y$, and $Z$ have not been summed at all. three partial states where $X$ and $Y$ have been summed, $Y$ and $Z$ have been summed, or $X$ and $Z$ have been summed, and one final state: $X$ and $Y$ and $Z$ have all been summed.
We articulate each of these states as interpretability hypotheses using high-level causal models that will be aligned to the \gpt\ model trained to perform addition. 

A layer of \gpt\ has not yet summed $X$, $Y$, and $Z$ if there are three components that separately represent $X$, $Y$, and $Z$. To localize each variable, we define models $\model^X, \model^Y, \model^Z$ with intermediate variable $P$ representing the processed input and $O$ representing the output:
\begin{align}
\text{$\model^X$:} \quad \mechanism{P}(X) &= X, & \mechanism{O}(P,Y,Z) &= P + Y + Z 
\end{align}

Similarly define $\model^{Y}$ and $\model^{Z}$. Also, we can define models where pairs of variables have been summed into an intermediate representation $P$, with the remaining variable to be added for the final sum in the output $O$:
\begin{align}
\text{$\model^{XY}$:} \quad \mechanism{P}(X,Y) &= X + Y, & \mechanism{O}(P,Z) &= P + Z 
\end{align}
Similarly define $\model^{YZ}$ and $\model^{XZ}$.
These three models represent intermediate computational states where exactly two of the input variables have been summed, while the third remains separate. Each model captures a different possible ordering of operations, reflecting different paths through the computation.
Lastly, define the model where all three variables are summed at the same time:
\begin{align}
\text{$\model^{XYZ}$:} \quad \mechanism{P}(X,Y,Z) &= X + Y + Z, & \mechanism{O}(P) &= P
\end{align}
This model represents the final computational state where all three numbers have been added together. The intermediate variable $P$ holds the complete sum, and the output mechanism simply returns this value unchanged.

\begin{figure}[t]
\centering
\includegraphics[width=0.75\textwidth]{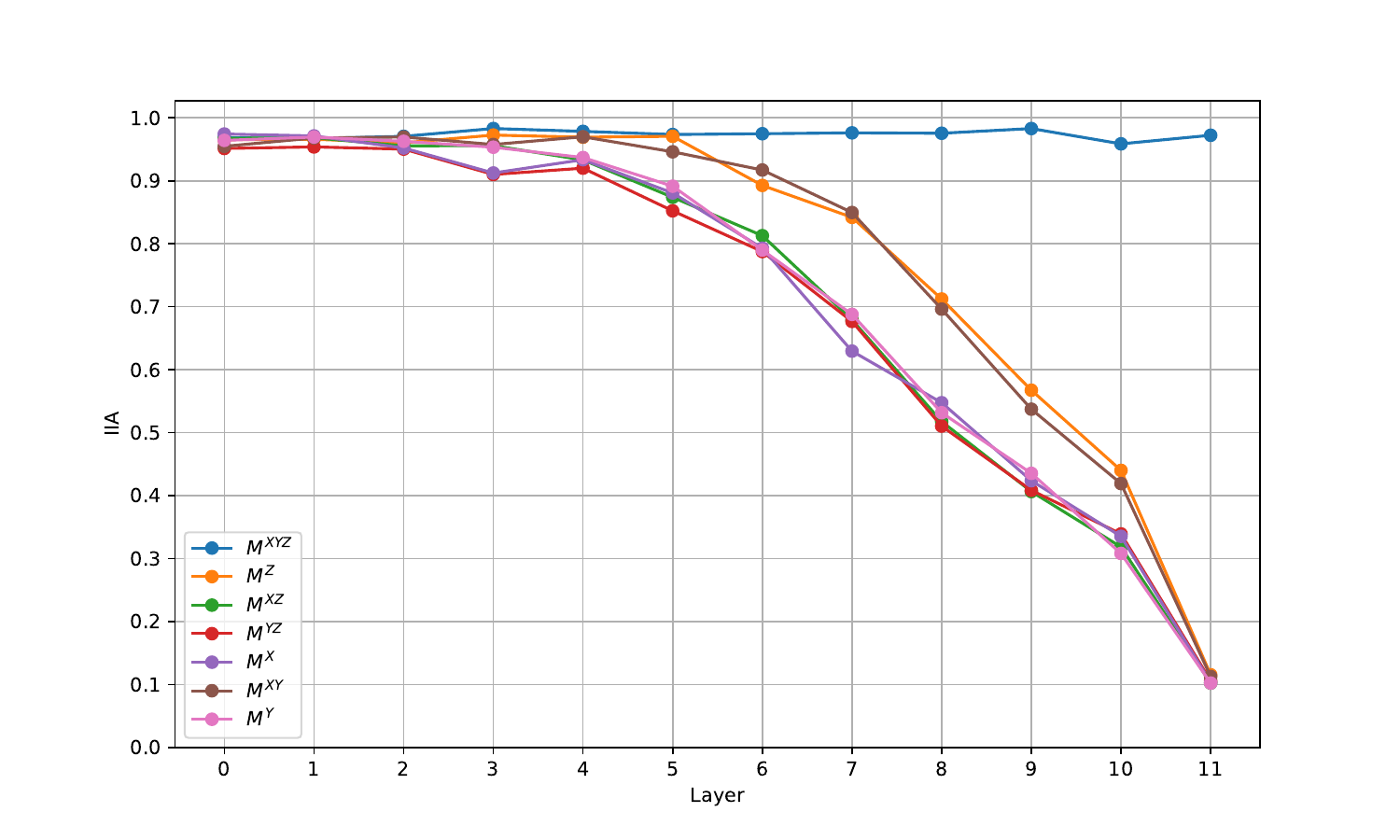}
\caption{Interchange Intervention Accuracy (IIA) for the intermediate variables $P$ of different high-level causal models across GPT layers when using a 256-dimensional alignment subspace. Early layers (1-4) show high accuracy for models representing separate variables ($\model^X$, $\model^Y$, $\model^Z$), indicating no summation has occurred. The complete summation model ($\model^{XYZ}$) maintains perfect accuracy across all layers. From layers 5-11, there is a gradual transition where partial summation models ($\model^{XY}$, $\model^{YZ}$, $\model^{XZ}$) show intermediate accuracy, suggesting the network does not transition between computational states in discrete steps.}
\label{fig:combined_empirical_plots}
\end{figure}

Each model captures a distinct hypothesis about the internal state of the network at a given layer, allowing us to track how the computation progresses through these states. Previous approaches would evaluate the degree to which a layer of \gpt\ adheres to each hypothesis.

\subsection{Distributed Alignment Search Results By Layer} For each causal model, we localize the intermediate variable $P$ to a $k$-dimensional linear subspace of the full residual stream of \gpt\ at a fixed layer $L$, i.e., the output of the $L$th transformer block with in $\mathbb{R}^{N \times d}$ where $N=6$ is the number of tokens (\textit{X+Y+Z=}) and $d=768$ is the model dimensionality. We use the implementation of distributed alignment search (DAS) from the \texttt{pyvene} library \citep{wu-etal-2024-pyvene}. We explore different values $k \in \{64, 128, 256\}$, and find stable results across these three settings. More information about the hyperparameters used can be found in Appendix \ref{appendix:training_intervenable_models}.
The results for $k = 256$ in terms of the Interchange Intervention Accuracy (IIA) for each of the 12 layers are shown in Figure~\ref{fig:combined_empirical_plots}. 

\paragraph{The Partial State $\model^{XY}$ is Most Accurate.} $\model^{XY}$ is more accurate for IIA than $\model^{YZ}$ and $\model^{YZ}$; the asymmetry in this solution is likely due to the causal attention of \gpt, i.e., the model can look at $X$, but not $Z$, when it processes $Y$.

\paragraph{Early and Late Layers are in Stable States.} In early layers of \gpt , the model is in a state where no summation has occurred. The models $\model^{X}$, $\model^{Y}$, and $\model^{Z}$ can each have their intermediate variable $P$ localized to the output of the transformer block up through layer 4 of \gpt . At layer 11, \gpt\ has completely summed the three numbers and no model can have its intermediate variable localized other than $\model^{XYZ}$. The model $\model^{XYZ}$ is perfectly accurate at all layers, which tells us that a $256$ dimensional subspace of the $5 \times 768$ dimension residual stream is sufficient to mediate the causal effect of inputs on outputs. 

\paragraph{For Many Intermediate Layers, the Model Partially Represents Multiple Quantities.} Between layers 4 and 11, there is a gradual transition.  The $\model^{XY}$ and $\model^{Z}$ are more accurate in terms of IIA than the other models, but everything other than $\model^{XYZ}$ (the trivial model) is only a partially accurate representation of what is going on at these layers. 
Based on these findings, we know that for each layer the model represents different quantities for different inputs. Our proposal is then to combine multiple causal models in order to form a new model that activates different causal processes based on the input provided to the model. This will allow us to construct a more faithful description of the network.

\section{Combining High-Level Abstract Models For More Faithful Abstractions of LLMs }
The approximate abstraction results in the previous section with partial faithfulness are difficult to interpret. If 70\% of interchange interventions are successful for a given alignment, does it make sense to think of the analysis as mostly successful, or could the high-level model be a completely misleading explanation of what is going on in the network? Early work attempted to avoid this issue by identifying the largest subset of the input space for which abstraction relation holds perfectly \citep{geiger-etal-2020-neural, geiger2021causal}. This specific evaluation was too rigid, as a single failed interchange intervention results in total failure. However, the general approach of adding nuance to the hypothesis in order to increase accuracy is promising.

In this work, we propose a flexible framework for combining and weakening interpretability hypotheses. The core idea is that subsets of model inputs are assigned to different high-level models which describe different points of a computational process. These high-level models are then \textit{combined} to form a more nuanced hypothesis that reflects the fact that the low-level network does not discretely transition between stages of computation, with sharp steps.

\begin{definition}[Combined Causal Models]
Let $\model^1, \dots, \model^k$ be causal models with identical input variables $\inputvars$ and output variables $\outputvars$, but distinct intermediate variables $\mathbf{V}_1, \dots, \mathbf{V}_k$, respectively. Define the combined $\model^*$ of causal models $\model^1, \dots, \model^k$ with input space partition $\Delta_1, \dots\Delta_k$ as follows. The intermediate variables of $\model^*$ are the union of all intermediate variables $\mathbf{V}^* = \bigcup^k_{j=1}\mathbf{V}_j$ and the mechanism for $X \in \mathbf{V}^*$ is 
\[\mathcal{F}^*_X(\mathbf{v}^*) = \begin{cases}
 \mathcal{F}_X^j(\project{\mathbf{v}^*}{\mathbf{V}_j}) & \mathrm{if } \ \project{\mathbf{v}^*}{\inputvars} \in \Delta_j\\
 \emptyset & \text{otherwise},\\
\end{cases}\]
where $\mathbf{v}^* \in \values{\mathbf{V}^*}$.
The mechanisms of the combined causal model $\mathcal{M}^*$ are a piecewise combination of the mechanisms from the uncombined models $\model^1, \dots, \model^k$. If an input $\mathbf{x} \in \values{\inputvars}$ is in partition $\Delta_j$, then the variables $\mathbf{V}_j$ are activated and the variables from other models are set to a null value $\emptyset$. 
\end{definition}

Let us demonstrate the combination of causal models with a concrete example from our arithmetic task. Consider combining three models: $\model^{XY}$ (sum $X,Y$ first), $\model^{YZ}$ (sum $Y,Z$ first), and $\model^{XYZ}$ (direct sum), with input space partitioned by the magnitude of $X$: $\Delta_1 = \{(x,y,z) : x \leq 3\}$, $\Delta_2 = \{(x,y,z) : 3 < x \leq 6\}$, and $\Delta_3 = \{(x,y,z) : x > 6\}$. The intermediate variables are $P_{XY}$, $P_{YZ}$, and $P_{XYZ}$ with mechanisms defined as follows:
$$\mechanism{P_{XY}}(x,y) = \begin{cases} x + y & \text{if } (x,y,z) \in \Delta_1 \\ \emptyset & \text{otherwise} \end{cases} \qquad \mechanism{P_{YZ}}(y,z) = \begin{cases} y + z & \text{if } (x,y,z) \in \Delta_2 \\ \emptyset & \text{otherwise} \end{cases}$$
$$\mechanism{P_{XYZ}}(x,y,z) = \begin{cases} x + y + z & \text{if } (x,y,z) \in \Delta_3 \\ \emptyset & \text{otherwise} \end{cases}$$
The output mechanism combines these intermediate computations:
$$\mechanism{O}(x,y,z,p_{xy},p_{yz},p_{xyz}) = \begin{cases} p_{xy} + z & \text{if } (x,y,z) \in \Delta_1 \\ x + p_{yz} & \text{if } (x,y,z) \in \Delta_2 \\ p_{xyz} & \text{if } (x,y,z) \in \Delta_3 \end{cases}$$
This combined model expresses the hypothesis that the network employs different addition strategies based on the magnitude of the input variable $X$: step-by-step addition for small numbers ($X \leq 3$), a different sequential strategy for medium-sized numbers ($3 < X \leq 6$), and direct computation for large numbers ($X > 6$).

\begin{figure}[t]
    \centering
    \subfigure[Comparison of X and Y term combinations]{
        \centering
        \includegraphics[width=0.48\linewidth]{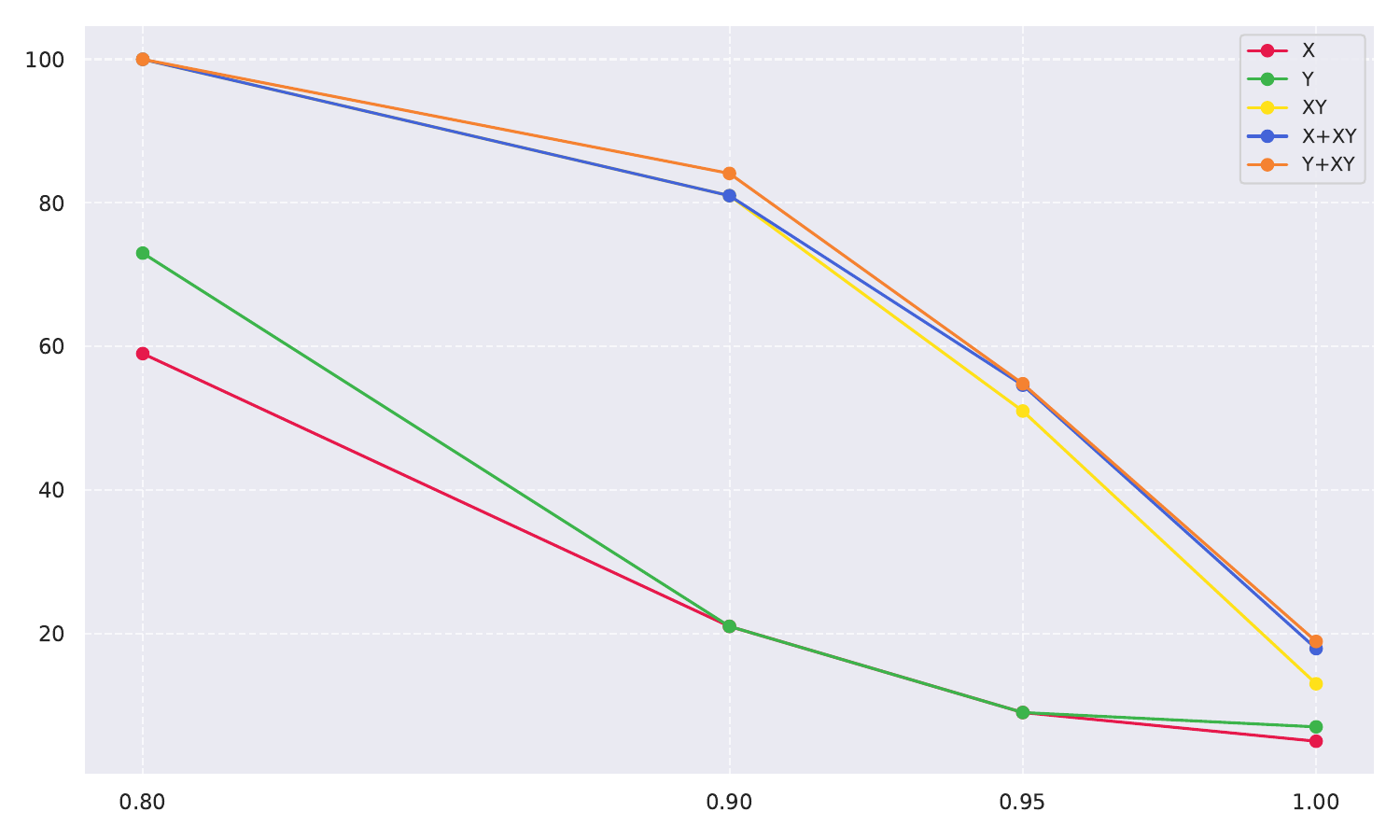}
        \label{fig:XY+}
    }
    \hfill
    \subfigure[Comparison of Y and Z term combinations]{
        \centering
        \includegraphics[width=0.48\linewidth]{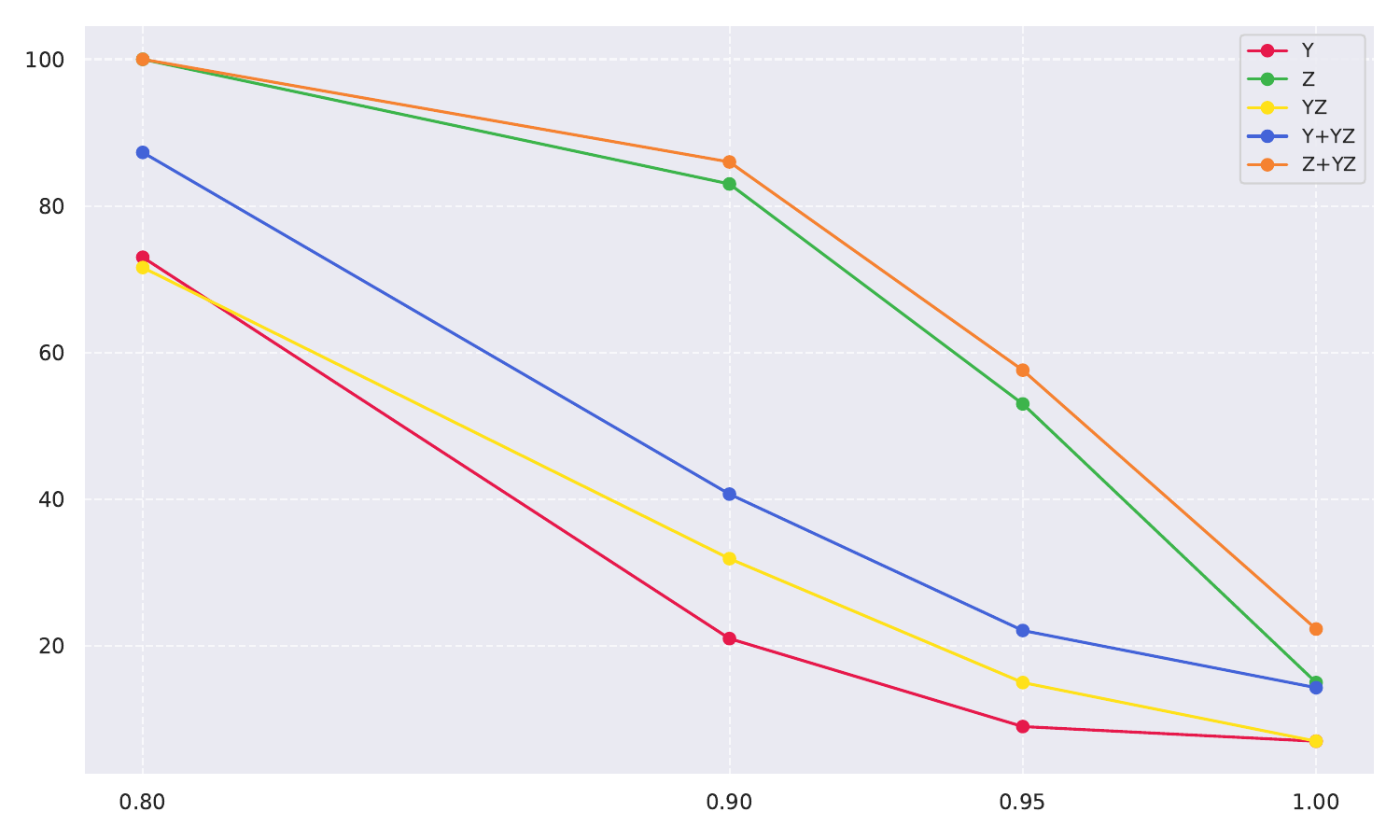}
        \label{fig:YZ+}
    }
    
    \subfigure[Comparison of X and Z term combinations]{
        \centering
        \includegraphics[width=0.48\linewidth]{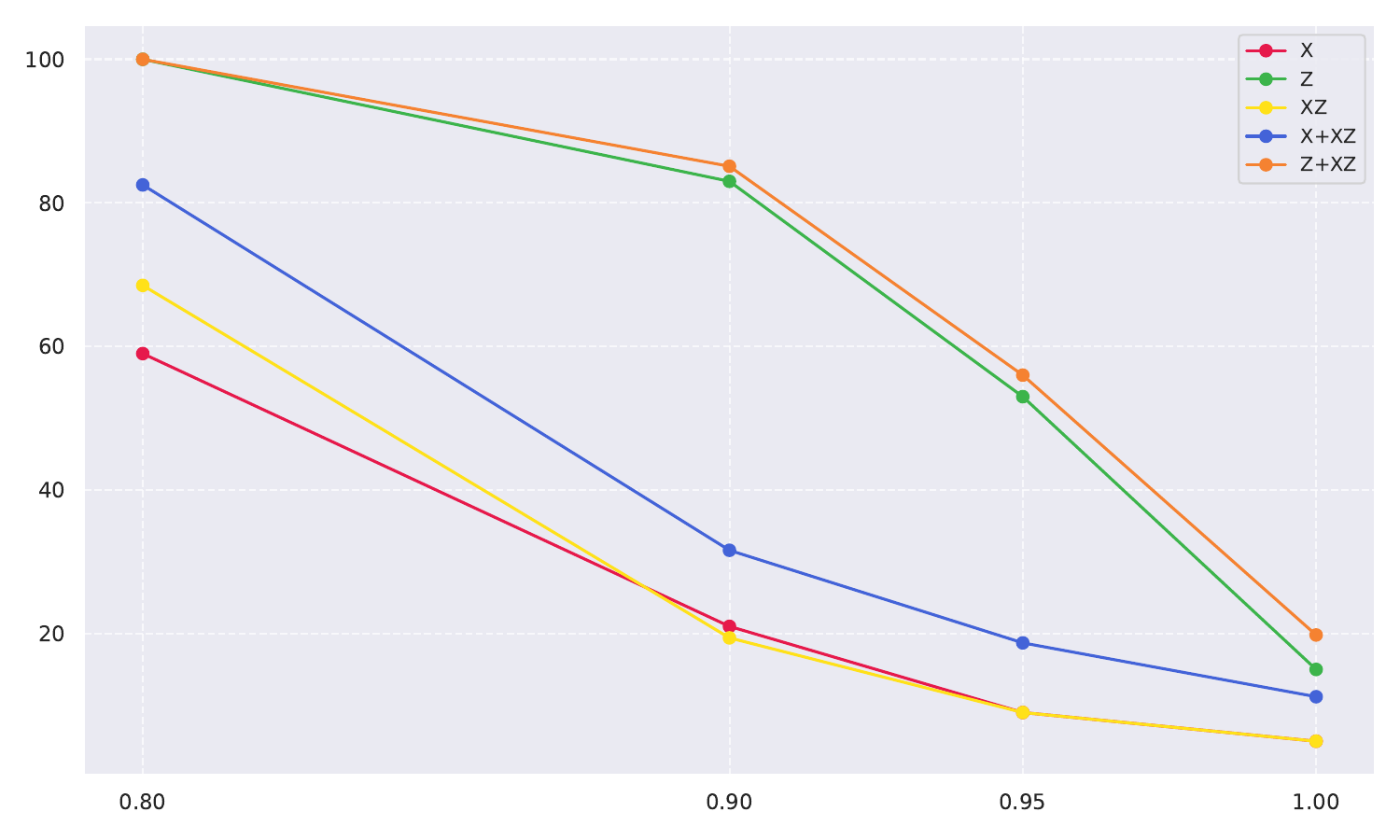}
        \label{fig:XZ+}
    }
    \hfill
    \subfigure[Overall comparison of all models]{
        \centering
        \includegraphics[width=0.48\linewidth]{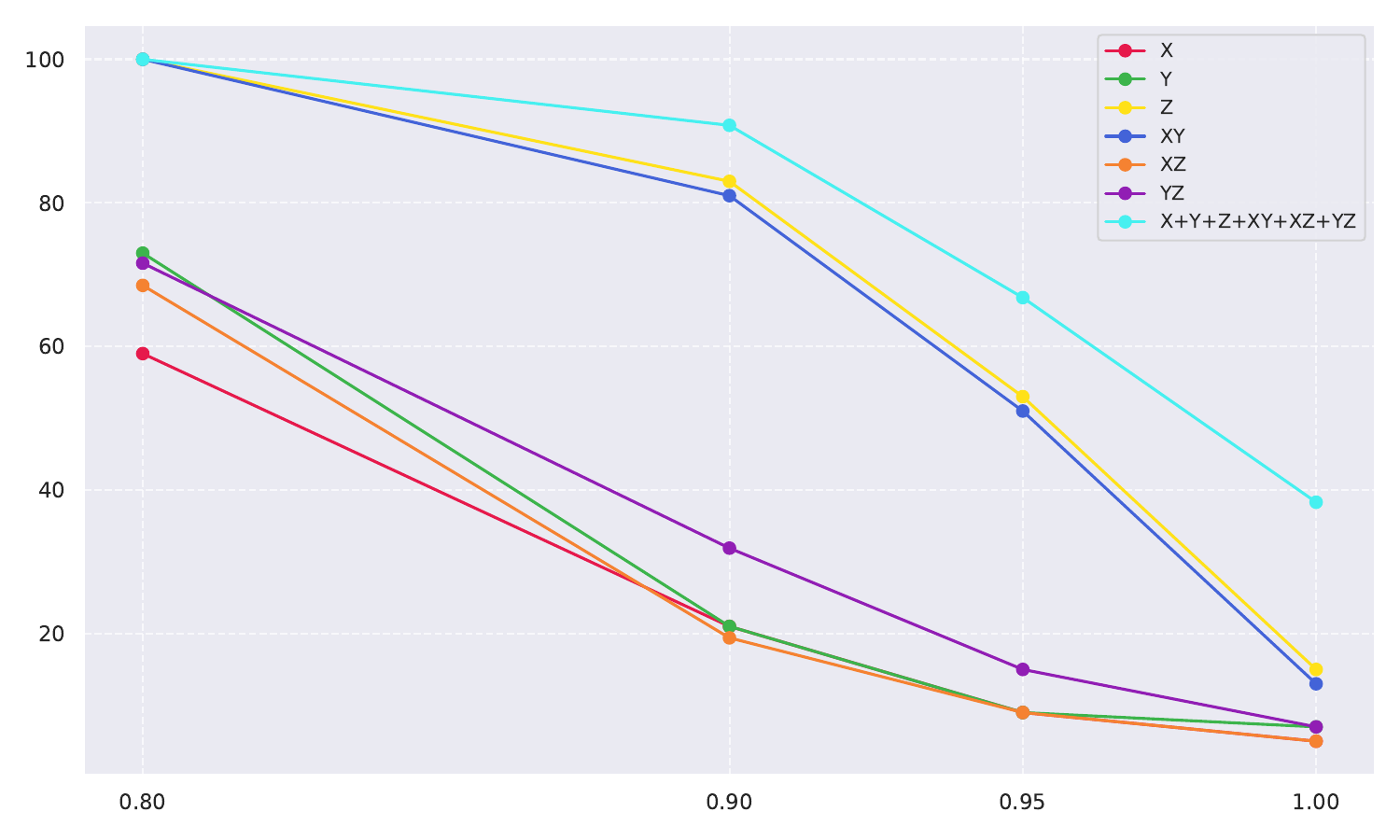}
        \label{fig:all}
    }
    \caption{Analysis of layer 7 causal models in fine-tuned GPT model on arithmetic tasks. The x-axis quantifies faithfulness with the interchange intervention accuracy achieved for the hypothesis. The y-axis quantifies strength as the proportion of inputs assigned to a non-trivial high-level causal model. Results compare the performance of individual causal models versus their combinations across different faithfulness thresholds. Combined models demonstrate stronger hypotheses at high faithfulness levels (0.9-1.0), while all models converge in performance at lower faithfulness thresholds (0.8).}
    \label{fig:arithcombine}
\end{figure}
\paragraph{The Faithfulness-Strength Trade-Off.}
The approach of combining models allows us to precisely quantify the trade-off between the strength of an interpretability hypothesis and its faithfulness to the underlying neural network. A stronger hypothesis makes more specific claims about the computational process by including more inputs in the partition cells for models with internal structure. At one extreme, we can assign all inputs to the model that makes no claims about intermediate computation (in the arithmetic task $\model^{XYZ}$), achieving perfect faithfulness but providing no insight into the network's operation. We will refer to this model as the \emph{trivial model}, since it is trivially perfectly accurate and, at the same time, still completely a black box. At the other extreme, we can assign all inputs to a single causal model, making a strong claim at the risk of inaccuracies. We will evaluate our proposed combined models on their ability to provide hypotheses that are both stronger and more faithful than any of the original causal models. 

Faithfulness is quantified as interchange intervention accuracy (See Section~\ref{sec:back}), i.e., degree to which neural network features play the same causal role as aligned high-level variables. We quantify strength as the proportion of inputs not assigned to a trivial model.
\begin{definition}[Model Strength]
Let $\model^*$ be a combined model of causal models $\model^1, \dots, \model^k$ with input space partition $\Delta_1, \dots, \Delta_k$. Without loss of generality, let $\model^k$ be the trivial model. The \emph{strength} of $\model^*$ is the proportion of inputs not assigned to the trivial model $1 - \frac{|\Delta_k|}{|\values{\inputvars}|}$.
\end{definition}

\paragraph{Aligning Combined Models.} In Section~\ref{sec:arith}, we described experiments where the intermediate variable from each causal model is aligned with a low dimensional linear subspace of the transformer residual stream. To align the variables in a combined model, we simply take the alignment learned for each of the original models.

\paragraph{Evaluation Graphs.} We need to assign as many inputs to the partition cells $\Delta_1, \dots, \Delta_{k-1}$ while remaining faithful to the low-level neural network. We precompute the interchange intervention accuracies for each model $\model^1, \dots \model^{k-1}$ in the form of \textit{evaluation graphs} $G^1, G^{k-1}$. In this weighted graph, the nodes are inputs, meaning there are $1000$ nodes for every possible $(x,y,z) \in \values{\inputvars}$.  For inputs $(x,y,z)$ and $(x',y',z')$, there are two interchange interventions depending on which input is the base and which is the source. The edge in $G^j$ is weighted 0, 0.5, or 1 according to the intervention interchange accuracy for model $\model^j$ on the two inputs.
$\model^j$ is a perfect abstraction of a neural network when its corresponding evaluation graph $G^j$ is a complete graph where all edges have weight 1. The pseudocode for obtaining an evaluation graph and a step-by-step example is available in Appendix \ref{app:evaluation_graphs}.

\paragraph{Greedily Constructing Input Space Partitions.}
We find the input partitions for a combined causal model using a greedy approach. The algorithm takes evaluation graphs $\{G^1, ..., G^k\}$ as an input and aims to partition the input space $\mathcal{X}$ across a set of candidate models $\{\model^1, ..., \model^k\}$, ensuring each model meets a minimum faithfulness threshold $\lambda$ on its assigned inputs. 

The algorithm is as follows. First, for each candidate model $\model^j$, we greedily identify the largest possible set of currently unassigned inputs on which $\model^j$ achieves the faithfulness threshold $\lambda$. 
\begin{enumerate}
    \item The nodes of $G^j$ are sorted from highest degree to lowest degree.
    \item Nodes are added to a subgraph $S^j$ until the next node would result in an interchange intervention accuracy, i.e., the average edge weight of $S^j$, that exceeds $\lambda$.
    \item The best model is $\model^j$ where $j = \mathsf{argmax}_j(|S^j|)$ is assigned input space $\Delta_j$ corresponding to the nodes of subgraph $S^j$.
\end{enumerate}
Then, the nodes from $S^j$ are removed from the graph, and we repeat the process until either all inputs have been assigned or no remaining model can faithfully handle any of the unassigned inputs. The algorithm returns both the set of selected models and their corresponding input partitions. The pseudocode and a step-by-step example can be found in Appendix~\ref{app:input_space_partitioning}.

\begin{figure}
\centering 
\subfigure[Evaluate the inner logic first and then apply the outer operator.]{
\resizebox{0.4\textwidth}{!}{
\begin{tikzpicture}[node distance=1cm, auto]
\tikzstyle{every node}=[circle,fill=green!50!black,minimum size=20pt,inner sep=0pt]
  \node (X) at (0,0) {$X$};
  \node (Op2) [left of=X] {$Op_2$};
  \node (X') [above of=X] {$X'$};
  \node (Op1) [left of=Op2] {$Op_1$};
  \node (Y) at (3,0) {$Y$};
  \node (Op3) [left of=Y] {$Op_3$};
  \node (Y') [above of=Y] {$Y'$};
  \node (B) [right of=X] {$B$};
  \node (Q) at (1,2) {$Q$};
  \node (O) at (0,3) {$O$};
  \foreach \from/\to in {X/X', Op2/X', Y/Y', Op3/Y', X'/Q, Y'/Q, B/Q, Op1/O, Q/O}
    \draw[->] (\from) -- (\to);
\end{tikzpicture}
}}
\hfill
\subfigure[Use De Morgan's laws, i.e., an input $\lnot (\lnot X \land Y)$ is converted to $X \lor \lnot Y$.]{
\resizebox{0.4\textwidth}{!}{
\begin{tikzpicture}[node distance=1cm, auto]
\tikzstyle{every node}=[circle,fill=green!50!black,minimum size=20pt,inner sep=0pt]
    \node (X2) at (0,0) {$X$};
    \node (Op22) [left of=X2] {$Op_2$};
    \node (X2') [above of=X2] {$X'$};
    \node (Op12) [left of=Op22] {$Op_1$};
    \node (Y2) at (3,0) {$Y$};
    \node (Op32) [left of=Y2] {$Op_3$};
    \node (Y2') [above of=Y2] {$Y'$};
    \node (B2) [right of=X2] {$B$};
    \node (V) at (-1,2) {$V$};
    \node (W) at (2,2) {$W$};
    \node (B') at (1,2) {$B'$};
    \node (O2) at (0,3) {$O$};
    \foreach \from/\to in {X2/X2', Op22/X2', Y2/Y2', Op32/Y2', X2'/V, Y2'/W, B2/B', Op12/B', B'/O2, Op12/V, Op12/W, V/O2, W/O2}
      \draw[->] (\from) -- (\to);
\end{tikzpicture}
}}
\caption{Two hypotheses for solving the boolean task.}
\label{fig:boolean_task_graphs}
\end{figure}
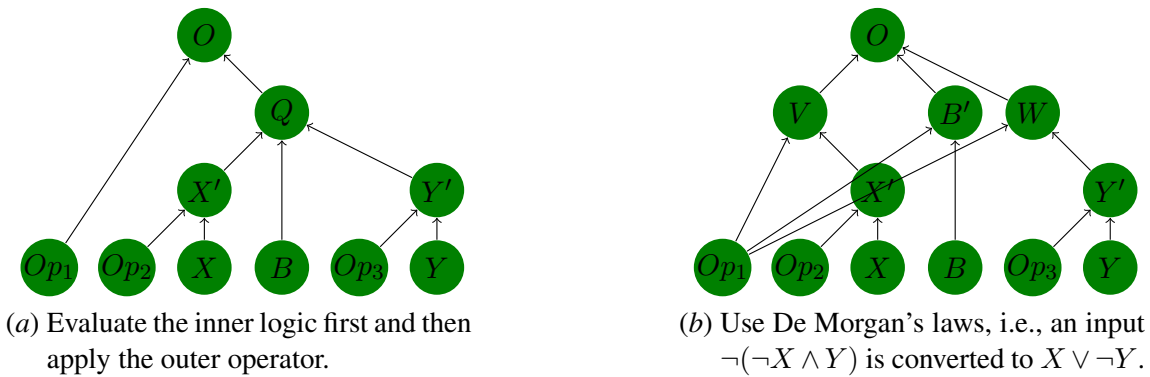

\section{Experimental results}
We demonstrate the effectiveness of our combination approach through experiments on \gpt\ fine-tuned on two toy tasks. While these tasks are deliberately simple to allow for clear analysis and validation, they serve as important proofs of concept. By showing that our method can successfully combine causal models on these controlled examples, we establish a foundation for applying these techniques to more complex architectures and real-world tasks. Our results consistently show that combining causal models leads to more robust and complete explanations of model behavior compared to analyzing individual causal models in isolation.

\subsection{Arithmetic Task Results}
We will now use our algorithm to analyze layer 7 from the \gpt\ model we fine-tuned on the arithmetic task, as described in Section~\ref{sec:arith}. For each faithfulness threshold (1, 0.95, 0.9, 0.8), we greedily assign inputs to the available causal models until the threshold is met. In Figure~\ref{fig:arithcombine}, we compare uncombined and combined causal models. 

\paragraph{Combined Models Provide Stronger Hypotheses at High Levels of Faithfulness.}
In Figure~\ref{fig:XY+}, we can see that the combination of $\mathcal{M}^{Y}$ or $\mathcal{M}^{X}$ with $\mathcal{M}^{Y + XY}$ is able to provide a stronger hypothesis than any of those models alone. Similarly in Figure~\ref{fig:YZ+} the combination of $\mathcal{M}^{Z}$ and $\mathcal{M}^{YZ}$ is stronger and in Figure~\ref{fig:XZ+} the combination of $\mathcal{M}^{Z}$ and $\mathcal{M}^{XZ}$ is stronger. In Figure~\ref{fig:all}, we can see that the full combination is the strongest overall for high-levels of faithfulness.

\begin{figure}[t]
\centering
\includegraphics[width=0.75\textwidth]{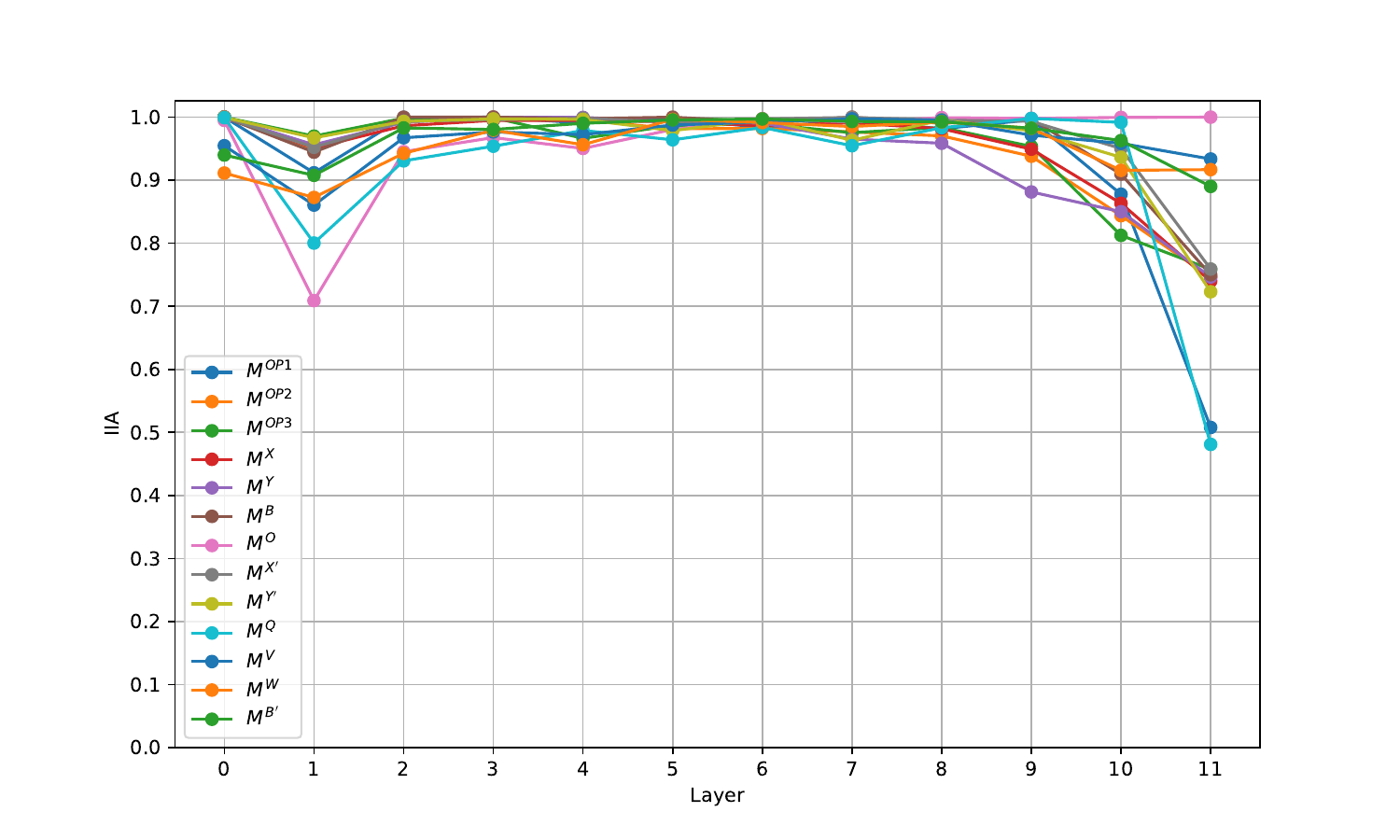}
\caption{Interchange Intervention Accuracy (IIA) for the intermediate variables $P$ of different high-level causal models of the boolean logic task across the layers of \gpt\ when searching within 256-dimensional subspaces of the neural representations.}
\label{fig:binary_combined_eval_plots}
\end{figure}

\begin{figure}[t]
    \centering
    \subfigure[$X'$ and components comparison]{
        \centering
        \includegraphics[width=0.48\linewidth]{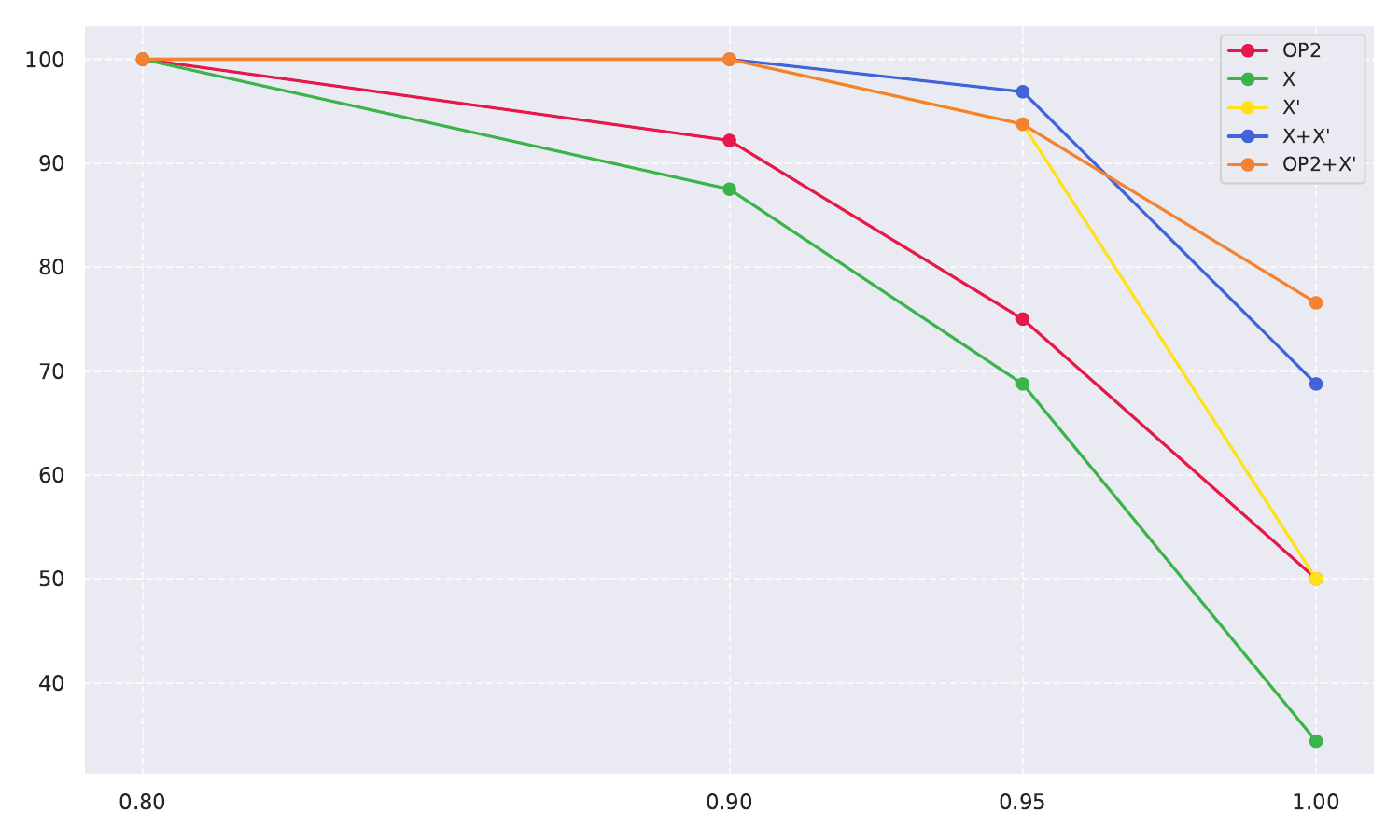}
        \label{fig:X'+}
    }
    \hfill
    \subfigure[$Y'$ and components comparison]{
        \centering
        \includegraphics[width=0.48\linewidth]{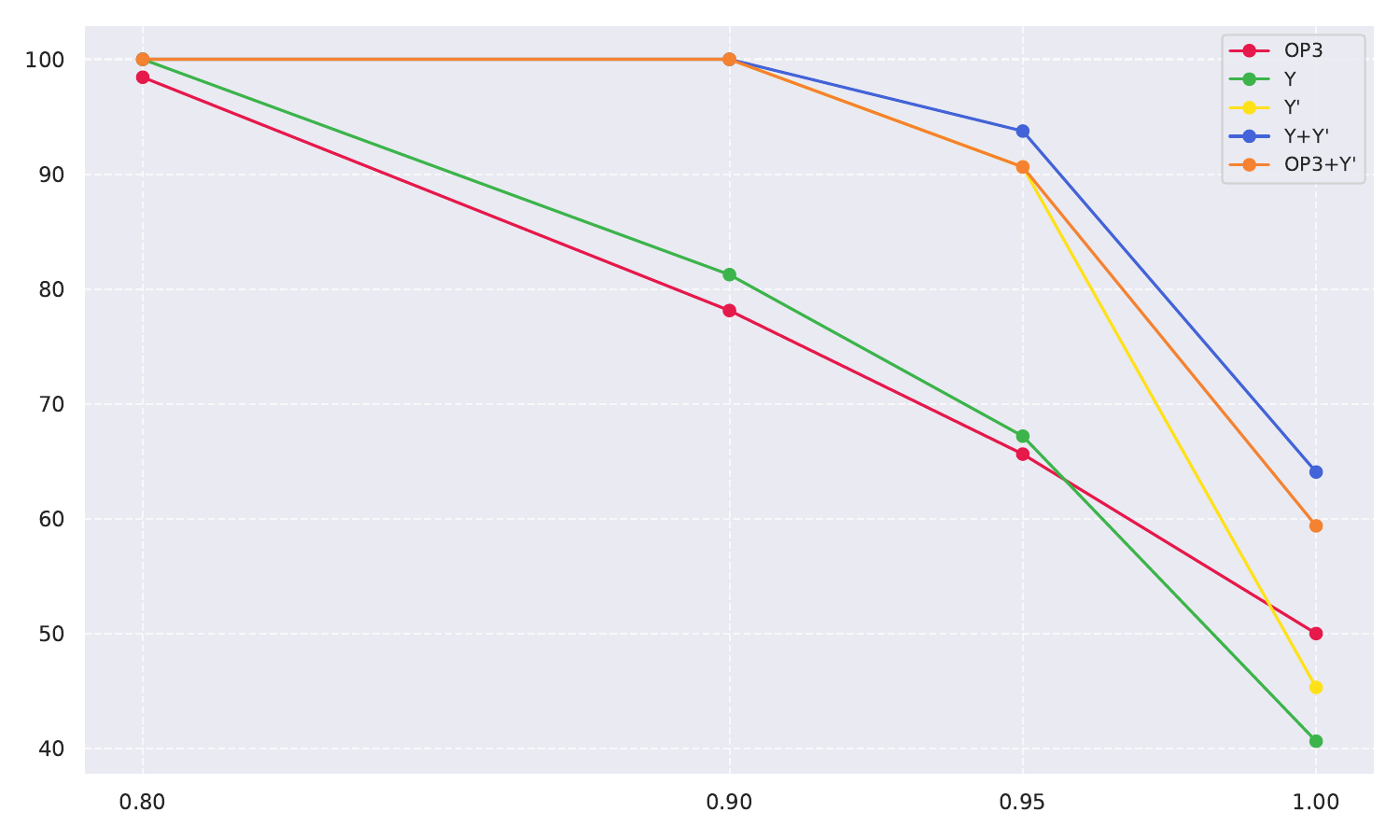}
        \label{fig:Y'+}
    }
    
    \subfigure[$V$ and components comparison]{
        \centering
        \includegraphics[width=0.48\linewidth]{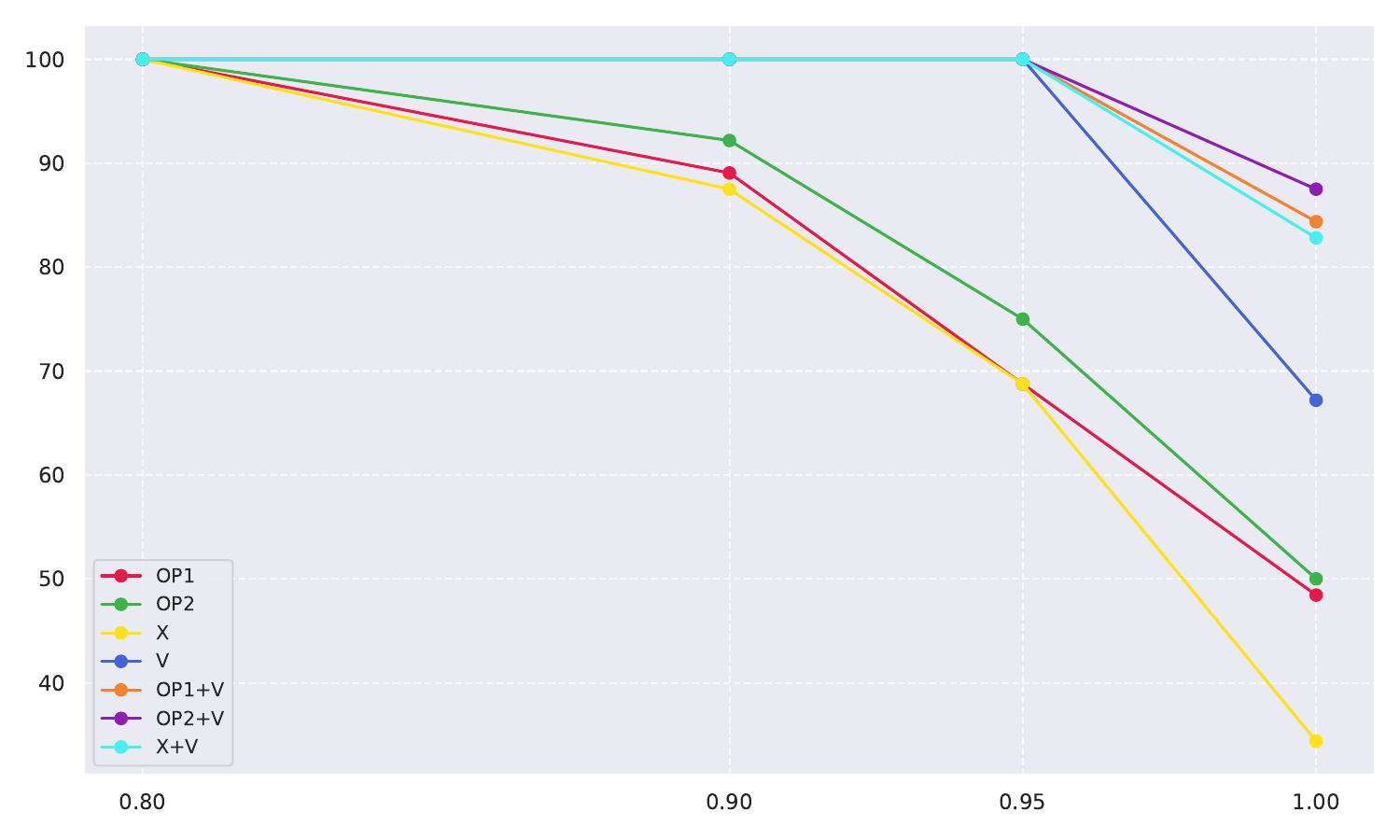}
        \label{fig:V+}
    }
    \hfill
    \subfigure[$W$ and components comparison]{
        \centering
        \includegraphics[width=0.48\linewidth]{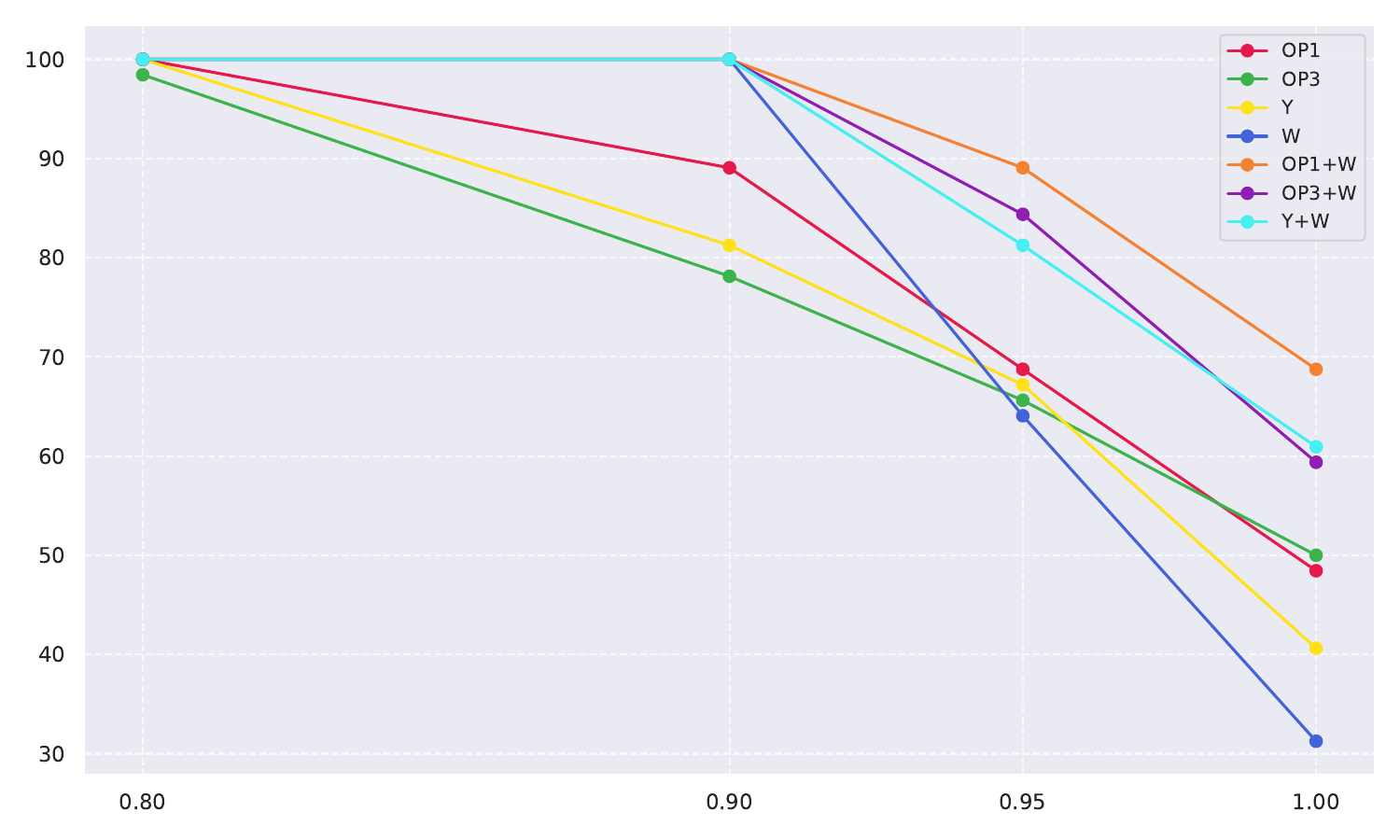}
        \label{fig:W+}
    }

    \subfigure[$Q$ and components comparison]{
        \centering
        \includegraphics[width=0.48\linewidth]{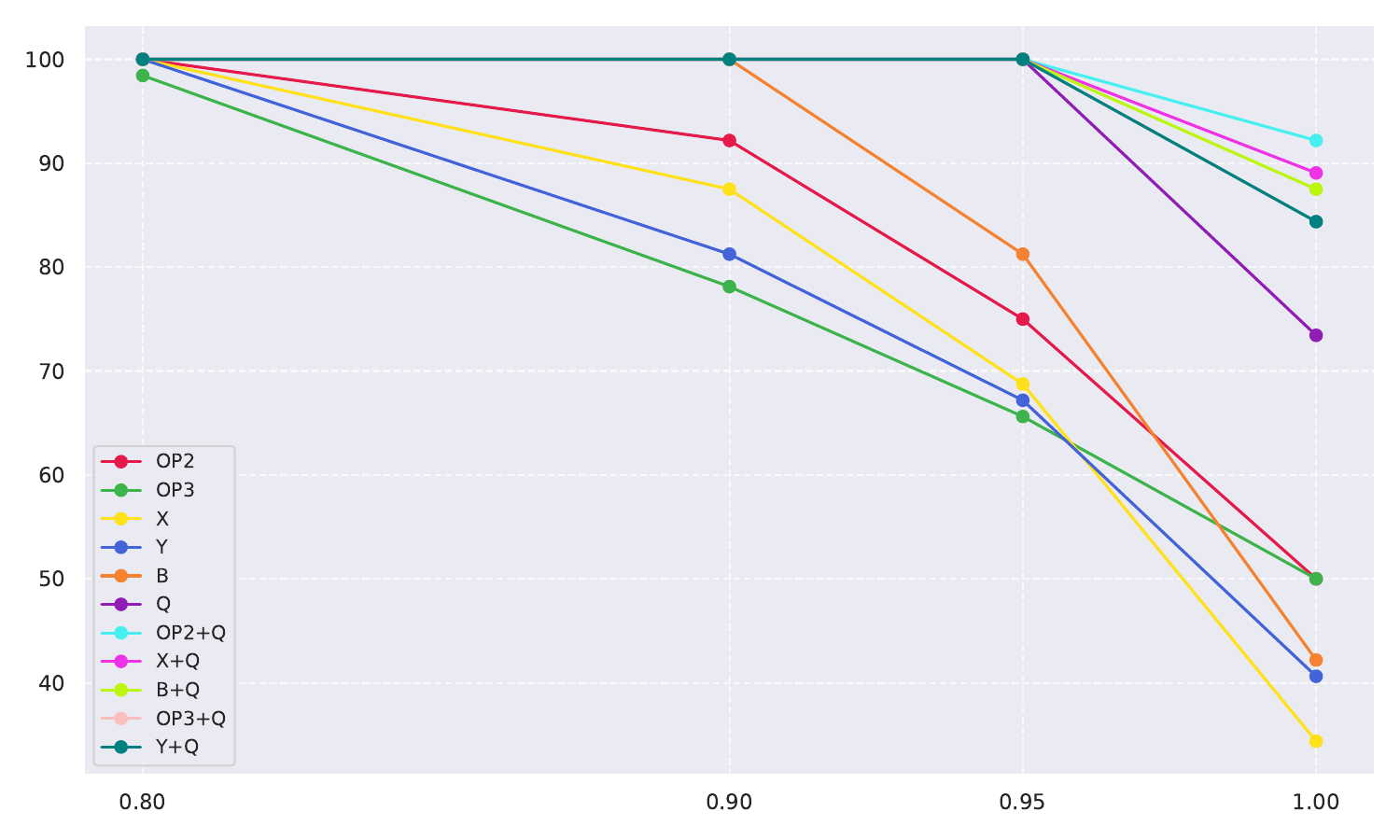}
        \label{fig:Q+}
    }
    \hfill
    \subfigure[$B'$ and components comparison]{
        \centering
        \includegraphics[width=0.48\linewidth]{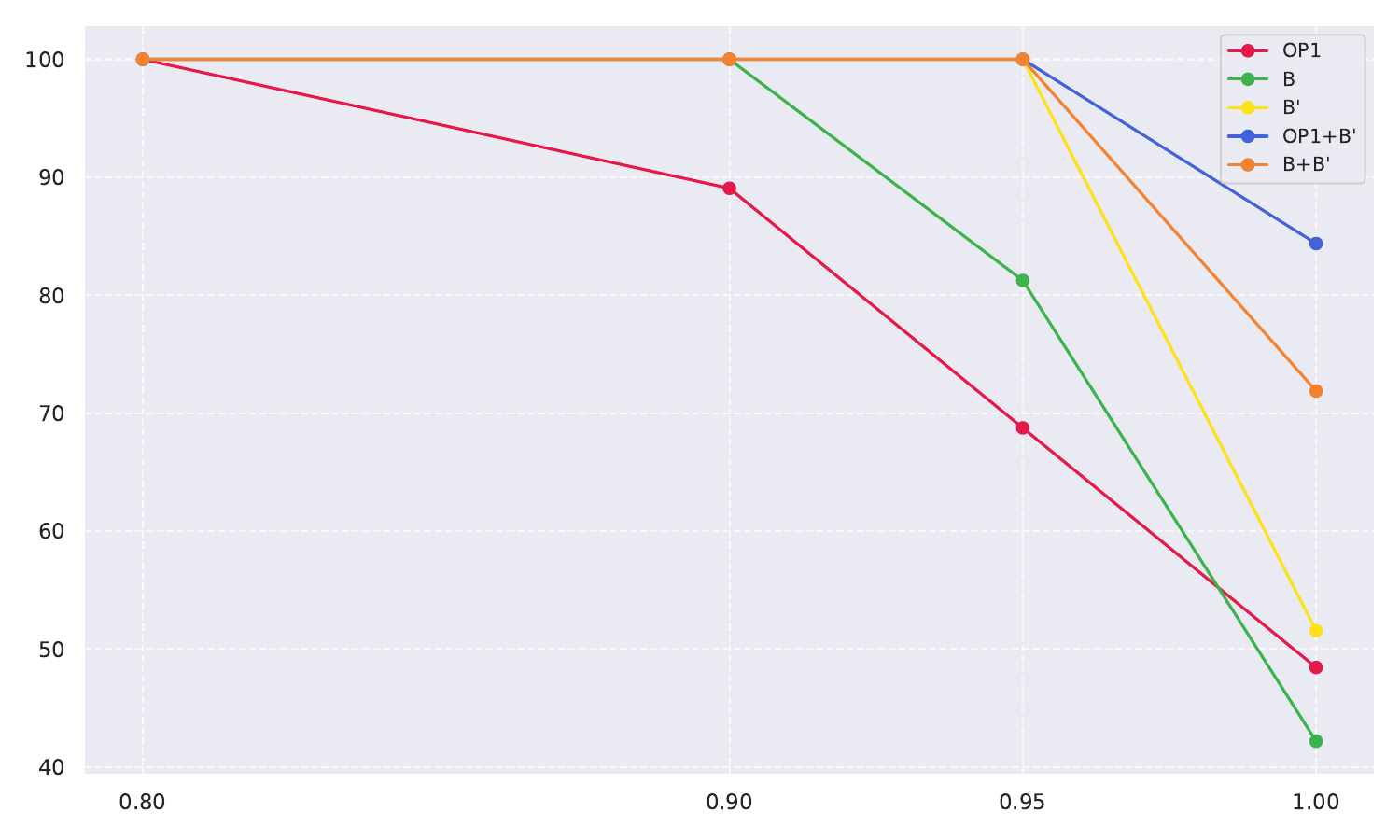}
        \label{fig:B'+}
    }
    
    \caption{Analysis of layer 10 causal models in fine-tuned GPT2-small model on boolean logic tasks. Results compare the performance of individual causal models versus their combinations across different faithfulness thresholds. Combined models demonstrate stronger hypotheses at high faithfulness levels (0.9-1.0).}
    \label{fig:binarycomb}
\end{figure}

\subsection{Boolean Logic Task}


Our second task is basic boolean logic with unary and binary operators. The task can be encoded into a prompt with the form $OP_1 \ (OP_2(X) \ B \ OP_3(Y))=$, where boolean variables $X,Y \in \{\texttt{True}, \texttt{False}\}$, and the unary operations $OP_1, OP_2, OP_3 \in \{\neg, \texttt{I}\}$, where $\neg$ represents the unary negation operator, while \texttt{I} is a unary identity operator. The $B$ is a boolean operator that can be either $\land$ (the AND operator) or $\lor$ (the OR operator). There are $64$ possible inputs.
We again fine-tune GPT2-small to perfectly predict the truth value of the input. Once again, it is unknown \textit{how} the model solves this task. However, we can come up with two intuitive solutions, and further split them into smaller states GPT2-small might encode within its neural representations.


\paragraph{Interpretability Hypotheses.} A direct solution is to evaluate the expressions $X':=OP_2(X)$ 
and $Y':=OP_3(Y)$, 
then evaluate $Q:=X' B Y'$, before finally evaluating $O:=OP_1(Q)$. This algorithm is shown in Figure \ref{fig:boolean_task_graphs}a.
The other approach is to use De Morgan's laws. In this case, the unary operator $OP_3$ is applied to $X'$, $B$, and $Y'$, to form the variable $V$, $B'$, and $W$, respectively, where $OP_3(\land) = \lor$ and $OP_3(\lor) = \land$. The causal model is shown in Figure \ref{fig:boolean_task_graphs}b. 
For each of these variables, we define a model with a single intermediate variable just as we did the arithmetic task. Explicit function definitions for each of the models refer to Appendix \ref{ap:models_boolean}.

\paragraph{Distributed Alignment Search Results By Layer.}

Figure \ref{fig:binary_combined_eval_plots} shows the evaluation of the boolean logic task in terms of IIA when searching for alignments within 256-dimensional linear subspaces for each of the 12 layers of GPT2-small. The prompt for the boolean task has a fixed size of 15 tokens, each encoded in a 768 dimensional space. This means that for the layer $L$, the Lth transformer block has a representation $\mathbb{R}^{15 \times 768}$. Observe that in layer 10, we begin to see a differentiation between the models, e.g., $\model^X$ has lower IIA compared to the ones that are dependent on more variables like $\model^{X'}$. Based on this, we focus our analysis on layer 10 of the model.

\paragraph{Combining Models For the Boolean Logic Task} 

We analyze layer 10 of GPT2-small fine-tuned on the boolean logic task. Similar to the previous experiment on the arithmetic task, for each faithfulness threshold (1,0.95, 0.9, 0.8). Figure \ref{fig:binarycomb} shows the comparison between combined variables and individual ones. We notice how across faithfulness levels for each of the combined variables models, the strength of a combined candidate model is higher than the result when considering each model individually. This stays consistent with the results from the previous experiments on the arithmetic task, even though the input space is much smaller (only 64 inputs compared to 1000 inputs). 
In Figure \ref{fig:X'+}, combining $\model^X$ or $\model^{OP_2}$ with $\model^{X'}$ resulting in $\model^{X+X'}$ or $\model^{OP_2+X'}$ can interpret a bigger input space compared to each of those models alone. For example, without combining the models, only 50\% or lower of data is interpretable with 100\% faithfulness, but when using a combined model, 75\% of data is interpretable with 100\% faithfulness. Also, in Figure \ref{fig:V+} we have that $\model^{OP2+V}$ abstracts 88\% of the input space with 100\% faithfulness, whereas the uncombined models only cover below 50\% of the input space. These trends can be observed across plots for different combined models. At 0.8 faithfulness, all models uncombined and combined models are equal at full strength.

\section{Conclusion.}
This work offers an approach to a fundamental challenge in mechanistic interpretability: the trade-off between hypothesis strength and faithfulness. By introducing a framework for combining causal models that activate different computational processes based on input, we enable more nuanced interpretations of neural computation. Our experiments on arithmetic and Boolean logic tasks with \gpt\ demonstrate that combined models can provide stronger interpretability hypotheses while maintaining high faithfulness compared to uncombined models, particularly at high faithfulness thresholds (0.9-1.0). While we focused on toy tasks for clear analysis and validation, this framework could be extended to more complex tasks, integrated with other mechanistic interpretability methods, and enhanced with more sophisticated optimization objectives for constructing input space partitions.

\clearpage

\acks{We thank SURF for the support in using the National Supercomputer Snellius. AG was supported by grants from
Open Philanthropy.}

\bibliography{bibentries, refs}

\clearpage
\appendix

\section{Algorithms} \label{app:algorithms}

There are two main algorithms our solution is based on: constructing evaluation graphs and partitioning inputs using these graphs. This section provides a detailed explanation of these approaches, including pseudocode and step-by-step examples.

\subsection{Evaluation Graphs Explained} \label{app:evaluation_graphs}

Each of the candidate causal models that we align with the LLM has a corresponding intervenable model trained to find these alignments, used to obtain its corresponding evaluation graph. The nodes in this graph correspond to the possible inputs of the task. For the sake of this example, we take only the case where in the arithmetic task, $X+Y+Z=$, we have $X, Y, Z \in \{1,2\}$. Therefore, there are $2^3 = 8$ possible inputs. The possible inputs and their corresponding outcomes are:

\begin{itemize}
    \item \{'X': 1, 'Y': 1, 'Z': 1\} with outcome 3.
    \item \{'X': 1, 'Y': 1, 'Z': 2\}, \{'X': 1, 'Y': 2, 'Z': 1\}, \{'X': 2, 'Y': 1, 'Z': 1\} with outcome 4
    \item \{'X': 1, 'Y': 2, 'Z': 2\}, \{'X': 2, 'Y': 1, 'Z': 2\}, \{'X': 2, 'Y': 2, 'Z': 1\} with outcome 5
    \item \{'X': 2, 'Y': 2, 'Z': 2\} with outcome 6
\end{itemize}

Therefore, we have 8 nodes in the evaluation graph. To construct the graph, we take each pair of nodes as input to a trained model for finding alignments between causal model $\model^k$ and a large language model. We call the trained model for finding alignments $I_{\model^k}$. We evaluate each input to be in turn the base and source for all the other nodes. For example, we evaluate both $I_{\model^k}(\texttt{base} = \{X: 1, Y: 1, Z: 1\}, \texttt{source} = \{X: 1, Y: 1, Z: 2\})$ and $I_{\model^k}(\texttt{base} = \{X: 1, Y: 1, Z: 2\}, \texttt{source} = \{X: 1, Y: 1, Z: 1\})$. Each evaluation yields a binary output (0 or 1), indicating the presence or absence of an alignment triggered by the counterfactual data obtained using that (base, source) unit. (Refer to Appendix \ref{sec:counterfactual_data} for details on the counterfactual dataset used to train/evaluate $I_{\model^k}$.)

The edge between two nodes is determined as follows:

\begin{itemize}
\item No edge if $I_{\model^k}(\texttt{base}, \texttt{source}) = I_{\model^k}(\texttt{source}, \texttt{base}) = 0$, indicating no detected relationship in either direction.
\item Edge weighted by 1 if $I_{\model^k}(\texttt{base}, \texttt{source}) = I_{\model^k}(\texttt{source}, \texttt{base}) = 1$, implying a strong alignment.
\item Edge weighted by 0.5 if $I_{\model^k}(\texttt{base}, \texttt{source}) \neq I_{\model^k}(\texttt{source}, \texttt{base})$, suggesting a directional alignment.
\end{itemize}

\begin{figure}[t]
\centering
\includegraphics[width=0.4\textwidth]{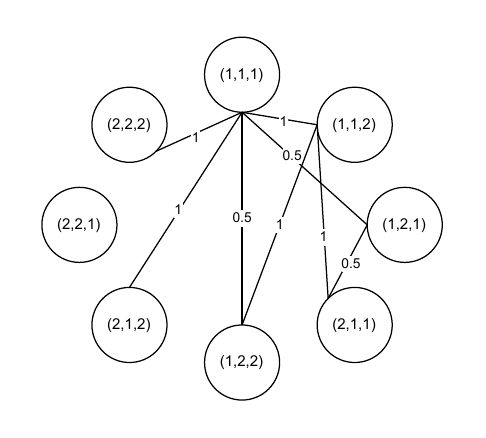}
\caption{Example of an evaluation graph for solving the arithmetic task when $X,Y,Z \in \{1,2\}$. There are 8 possible inputs, corresponding to the nodes in the graph. The lack of connections to node (2, 2, 1) indicates that a different, more appropriate intervenable model is likely required to understand how the LLM processes this particular input.}
\label{fig:eval_graph_example}
\end{figure}

Each graph has a corresponding IIA computed as:

\begin{equation}
    \texttt{IIA}_{G^k} = 2 *\frac{\sum\limits_{(i,j) \in E} w(i,j)}{|V| * (|V| - 1)} 
\end{equation}

where $E$ is the set of edges in graph $G^k$, $|V|$ represents the number of nodes and $w(i,j)$ represents the weight between nodes $i$ and $j$. 

A fully connected graph with all edges weighted by 1 signifies a perfect alignment between the causal model $\model^k$ and the LLM's behaviour. Figure \ref{fig:eval_graph_example} shows an example of how the graph in the arithmetic task could look like when $X, Y, Z \in \{1, 2\}$. Algorithm \ref{alg:graph_construction} provides the pseudocode for constructing these evaluation graphs.

\begin{algorithm}
\caption{Construction of an evaluation graph for a candidate model.}
\label{alg:graph_construction}

\KwIn{list of all possible inputs for a task V, trained alignment $I_{\model^k}$}
\KwOut{evaluation graph $G^k$}

$G^k$ $\gets$ $|V| \times |V|$ zero matrix

\For{$i \gets 0$ \textbf{to} $|V| - 1$}{
    \For{$j \gets i + 1$ \textbf{to} $|V| - 1$}{
      $base\_source \gets$ counterfactuals when V[i] is the base and V[j] is the source \\
      $source\_base \gets$ counterfactuals when V[i] is the source and V[j] is the base \\
      $G^k[i][j] = G^k[j][i] \gets$ evaluate $I_{\model^k}$ on $[base\_source, source\_base]$
    }
}
\end{algorithm}

\subsection{Input Space Partitioning Explained} \label{app:input_space_partitioning}

The core of our approach lies in partitioning the input space using the previously constructed evaluation graphs. This process aims to greedily maximize the portion of the input space explained within a defined faithfulness threshold, $\lambda$. In other words, the goal is to find a stronger hypothesis (combined model), where strength is defined by how little input space is left unassigned to an explanation.

\begin{figure}
\centering 
\subfigure[$G^i$]{
\resizebox{0.45\textwidth}{!}{
\includegraphics[width=0.4\textwidth]{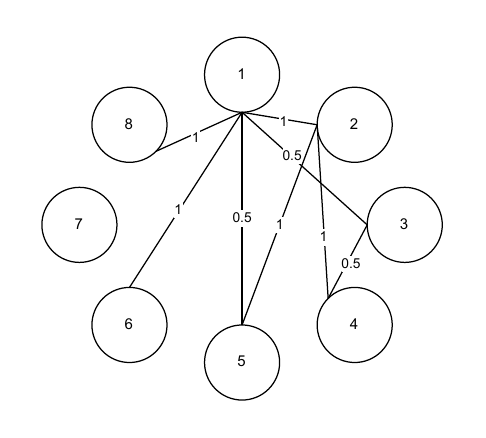}
}}
\hfill
\subfigure[$G^j$]{
\resizebox{0.45\textwidth}{!}{
\includegraphics[width=0.4\textwidth]{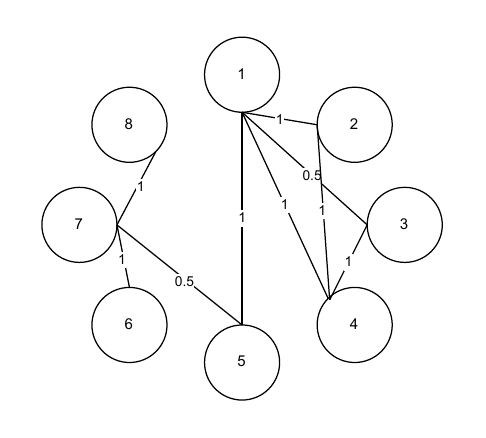}
}}
\caption{Example of two evaluation graphs, $G^i$ and $G^j$, constructed by evaluating $I_{\model^i}$ and $I_{\model^j}$, respectively.}
\label{fig:graphs_example}
\end{figure}

Figure \ref{fig:graphs_example} illustrates two evaluation graphs, $G^i$ and $G^j$. We use them as examples to demonstrate how we combine their corresponding causal models to form $\model^{i+j}$ when the faithfulness level is $\lambda = 0.4$. Here is a step-by-step breakdown:

\begin{itemize}
    \item We begin by choosing the graph with the highest IIA. Therefore, we start with $G^j$ which has an IIA of $\frac{8}{28}$.
    \item The nodes of $G^j$ are sorted in descending order based on their degree: 1,7,4,5,3,2,8,6.
    \item We greedily build a subgraph by iteratively adding nodes in the sorted order, ensuring the resulting subgraph maintains an IIA above the faithfulness threshold ($\lambda = 0.4$). This process results in the subgraph containing nodes 1, 7, 4, 5, and 2, with an IIA of 0.45.
    \item We move on to the next highest IIA graph, $G^i$, and order its nodes by degree in descending order: 1, 2, 5, 4, 3, 8, 6, 7.
    \item We exclude nodes already included in the subgraph from $G_j$, leaving the ordered list of nodes 3,8,6.
    \item There are no connections between the remaining nodes 3,8,6 in $G^i$. Therefore, adding these nodes would not enhance the explanation within the required faithfulness threshold.
    \item The portion of the input space that is left unassigned is handled by an output model. In this example, $\model^{i+j}$ explained 62\% of the input space with 45\% faithfulness.
\end{itemize}

This is just an example to showcase how partitioning the input space works. Algorithm \ref{alg:combined_algorithm} provides the pseudocode for implementing this input space partitioning based on evaluation graphs.

\begin{algorithm}
\caption{Greedy algorithm for selecting the inputs that correspond to each model.}
\SetAlgoLined
\KwIn{Faithfulness threshold $\lambda$, candidate models $\mathcal{M} = \{\model^1, ..., \model^k\}$, input space $\mathcal{X}$, evaluation graphs $\mathcal{G} = \{G^1, .., G^k\}$}
\KwOut{Set of selected models and their input partitions}
 assigned\_inputs, selected\_models $\gets \emptyset, \emptyset$\;\\
\While{$\mathcal{X} \setminus$ assigned\_inputs $\neq \emptyset$}{
    best\_model $\gets$ null\;\\
    best\_inputs $\gets \emptyset$\;\\
    \ForEach{$\model^j \in \mathcal{M}$}{
        current\_inputs $\gets \emptyset$\;\\
        $g^k$ $\gets$ $G^k(\mathcal{X} \setminus \text{assigned\_inputs})$\\
        $temp\_g^k \gets \emptyset$\\
        \ForEach{$x \in$ nodes sorted by degree in subgraph $g^k$}{
            $temp\_g^k$ $\gets$ $temp\_g^k \cup \{x\}$\\
            \If{$IIA_{temp\_g^k} \geq \lambda$}{
                current\_inputs $\gets$ current\_inputs $\cup \{x\}$\;
            }
        }
        \If{$|$current\_inputs$| > |$best\_inputs$|$}{
            best\_model $\gets \model^j$\;\\
            best\_inputs $\gets$ current\_inputs\;
        }
    }
    \If{best\_model = null}{
        \textbf{Break}\;
    }
    selected\_models $\gets$ selected\_models $\cup \{$best\_model$\}$\;\\
    assigned\_inputs $\gets$ assigned\_inputs $\cup$ best\_inputs\;
}
\Return (selected\_models, assigned\_inputs)
\label{alg:combined_algorithm}
\end{algorithm}

\section{LLM Selection and Fine-tuning} \label{sec:fine-tuning_gpt2}

The fine-tuning process was implemented using the Hugging Face Transformers library, which offers a comprehensive toolkit for working with pre-trained language models. 
Because for our arithmetic or boolean logic task the goal is to predict the correct output of the sum of three numbers or the correct evaluation of a logic expression, respectively, the language modelling capabilities of GPT-2 are not needed. Therefore, we employ the \texttt{GPT2ForSequenceClassification} class, a specialized variant of GPT-2 designed for sequence classification tasks. The fine-tuning process involved a selection of hyperparameters to optimize performance on the two tasks. The key hyperparameters and their chosen values are presented in Table \ref{tab:hyperparameters_gpt2}.

\begin{table}[h!]
\centering
\begin{tabular}{ll}
\toprule
\textbf{Hyperparameter} & \textbf{Value} \\
\midrule
Training Size & 2560 \\
Learning Rate & 2e-5 \\
Optimizer & AdamW \\
Batch Size & 32 \\
Epochs & 50 \\
\bottomrule
\end{tabular}
\caption{Hyperparameters for fine-tuning the GPT2 to the arithmetic and boolean logic task.}
\label{tab:hyperparameters_gpt2}
\end{table}

\section{Data Generation} \label{appendix:data}

The data which is needed throughout the pipeline of this research is of two types: factual, which is observational data used for fine-tuning the LLM, and counterfactual, which is interventional data used to find alignments. Utilizing black box models, such as LLMs, has the advantage of easily generating counterfactual data. Black boxes are closed systems, where the internal mechanisms are abstracted away. By simply manipulating the input parameters and observing the corresponding outputs, one can explore different hypothetical scenarios, effectively simulating counterfactual conditions. Since they are closed systems, the output can be simultaneously analysed in factual and counterfactual contexts.

\subsection{Factual Datasets}

To obtain a \textit{factual dataset}, one must construct a standard input-taking prompt. The inputs must match the input variables in the causal models that one wishes to align. All the causal models which solve the same task share the same factual dataset. For the arithmetic task that we employ, we generate factual data using the prompt below, where the inputs $X, Y, Z$ are numerical values between 1 and 10:

\begin{tcolorbox}
\begin{lstlisting}
X+Y+Z=
\end{lstlisting}
\label{prompt1}
\end{tcolorbox}

For the arithmetic task, we finetune the GPT2-small generating prompts of the form, where the inputs are defined as in the main paper:

\begin{tcolorbox}
\begin{lstlisting}
OP1(OP_2(X) B OP_3(Y))=
\end{lstlisting}
\label{prompt2}
\end{tcolorbox}

\subsection{Counterfactual Datasets} \label{sec:counterfactual_data}

Generating the counterfactual data requires two generated inputs: a base and a source. Depending on the intervenable variable, the counterfactual output is obtained by replacing the base input under that intervenable variable with the source sample under the same intervenable input.

\begin{figure}[h]  
    \centering
    \includegraphics[width=0.5\linewidth]{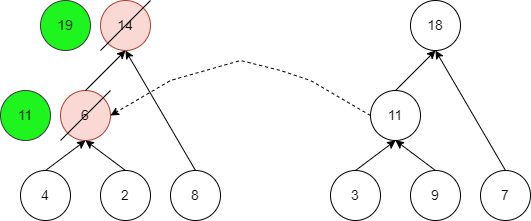}
    \caption{Example of obtaining counterfactual data where the base is $\{X: 4, Y: 2, Z: 8\}$ with intermediate variable $P = X + Y = 6$ and output $O = P + Z = 14$ , and the source is $\{X: 3, Y: 9, Z: 7\}$ with intermediate variable $P = X + Y = 11$ and output $O = P + Z = 18$. 
    }
    \label{fig:counterfactual_data}
\end{figure}

For example, in Figure~\ref{fig:counterfactual_data}, one can visualize how such counterfactual data is obtained for the arithmetic task. The first level in the trees represents the input, everything that is in between the first level and the last represents the intervenable variables, and the last level represents the outcome. In the example graphs, the base is the tuple (4, 2, 8) with outcome 14, and the source is the tuple (3, 9, 7). Applying an \textit{interchange intervention (II)} between the source and the base at the intervenable variable means replacing the outcome of the intervenable variable of the base, which is 6, with the outcome of the intervenable variable of the source, which is 11. Therefore, the counterfactual outcome becomes 19.

\subsection{Generation of Graph Nodes} \label{sec:nodes_graph_generation}

One essential part of our analysis is to construct the graph \(G = (V, E, w)\). In this section, we explain how the set of vertices, $V$, is obtained. The input to the arithmetic task is constrained to the integer interval [1, 10]. Each node within the graph is represented by a tuple of three such integers. We systematically generate all possible permutations of three numbers within this interval, associating each tuple with a distinct node. Given that the number of arrangements of three elements from a set of ten is 10 * 10 * 10 = 1000, the graph consists of 1000 nodes. The set of nodes $V$ in graph $G$ is formally defined in Equation \ref{eq:permutations}.

\begin{equation}
\label{eq:permutations}
    V := \{ (X, Y, Z) \mid X, Y, Z \in {1, 2, ..., 10}\}
\end{equation}

Similarly, for the boolean logic task, we have two possible values for each of the 6 inputs $OP_1$, $OP_2$, $X$, $B$, $OP_3$, $Y$, and every combination of their values results in $2 * 2 * 2 * 2 * 2 * 2 = 64$ possible inputs.

\section{Training Intervenable Models} \label{appendix:training_intervenable_models}

Given the structure of our arithmetic task, interventions are performed on up to 6 tokens simultaneously, corresponding to the maximum number of tokens in the prompt $'X+Y+Z='$.
Besides the explored subspace dimensions, $low\_rank\_dimension  \in \{64, 128, 256\}$, for causal models $M_X$ and $M_O$ we also test for lower dimensions such as $low\_rank\_dimension \in \{4, 8, 16, 32\}$. By varying the subspace dimension, we can assess the trade-off between computational efficiency and the ability to capture nuanced representations of the arithmetic operations within these lower-dimensional spaces. In total, there are 216 trained intervenable models for each of the 6 causal models in the arithmetic task, because on each of the 12 layers, for each of the three low-rank dimensions $\{64, 128, 256\}$, an alignment is searched for through training (216 trained intervenable models), and for two of the causal models, an additional of 96 models are trained, because we also search alignments in 4 additional low rank dimensions $\{4, 8, 16, 32\}$. For each of the tasks, the training hyperparameters are shown in Table \ref{tab:hyperparameters_intervenable}.

\begin{table}[h!]
\centering
\begin{tabular}{lll}
\toprule
\textbf{Hyperparameter} & \textbf{Arithmetic} & \textbf{Boolean} \\
\midrule
Training Size & 256000 & 4096 \\
Learning Rate & 0.01 & 0.01 \\
Batch Size & 1280 & 128\\
Epochs & 4 & 5\\
\bottomrule
\end{tabular}
\caption{Hyperparameters for training intervenable models for the arithmetic and boolean tasks.}
\label{tab:hyperparameters_intervenable}
\end{table}

\section{Models For The Boolean Task} \label{ap:models_boolean}

The models below represent the intermediate states GPT2-small could be in when solving the boolean logic task. The explicit function definitions are listed below.

\begin{center}
\tiny
\begin{align*}
\text{$\model^{X}$:} \quad \mechanism{P}(X) &= X, & \mechanism{O}(OP_1, OP_2, P, B, OP_3, Y) &= OP_1 \ (OP_2(P) \ B \ OP_3(Y)) \\
\text{$\model^{Y}$:} \quad \mechanism{P}(Y) &= Y, & \mechanism{O}(OP_1, OP_2, X, B, OP_3, P) &= OP_1 \ (OP_2(X) \ B \ OP_3(P)) \\
\text{$\model^{B}$:} \quad \mechanism{P}(B) &= B, & \mechanism{O}(OP_1, OP_2, X, P, OP_3, Y) &= OP_1 \ (OP_2(X) \ P \ OP_3(Y))
\\
\text{$\model^{OP_1}$:} \quad \mechanism{P}(OP_1) &= OP_1, & \mechanism{O}(P, OP_2, X, B, OP_3, Y) &= P \ (OP_2(X) \ B \ OP_3(Y)) \\
\text{$\model^{OP_2}$:} \quad \mechanism{P}(OP_2) &= OP_2, & \mechanism{O}(OP_1, P, X, B, OP_3, Y) &= OP_1 \ (P(X) \ B \ OP_3(Y)) \\
\text{$\model^{OP_3}$:} \quad \mechanism{P}(OP_3) &= OP_3, & \mechanism{O}(OP_1, OP_2, X, B, P, Y) &= OP_1 \ (OP_2(X) \ B \ P(Y)) 
\end{align*}

\begin{align*}
\text{$\model^{X'}$:} \quad \mechanism{P}(OP_2, X) &= OP_2(X), & \mechanism{O}(OP_1, P, B, OP_3, Y) &= OP_1 \ (P \ B \ OP_3(Y))
\\
\text{$\model^{Y'}$:} \quad \mechanism{P}(OP_3, Y) &= OP_3(Y), & \mechanism{O}(OP_1, OP_2, X, B, P) &= OP_1 \ (OP_2(X) \ B \ P)
\\
\text{$\model^{Q}$:} \quad \mechanism{P}(OP_2, X, B, OP_3, Y) &= OP_2(X) \ B \ OP_3(Y), & \mechanism{O}(OP_1, P) &= OP_1 \ (P)
\\
\text{$\model^{V}$:} \quad \mechanism{P}(OP_1, OP_2, X) &= OP_1(OP_2(X)), & \mechanism{O}(OP_1, P, B, OP_3, Y) &= P \ OP_1(B) \ OP_1(OP_3(Y))
\\
\text{$\model^{W}$:} \quad \mechanism{P}(OP_1, OP_3, Y) &= OP_1(OP_3(Y)), & \mechanism{O}(OP_1, OP_2, X, B, P) &= OP_1(OP_2(X)) \ OP_1(B) \ P
\\
\text{$\model^{B'}$:} \quad \mechanism{P}(OP_1, B) &= OP_1(B), & \mechanism{O}(OP_1, OP_2, X, P, OP_3, Y) &= OP_1(OP_2(X)) \ P \ OP_1(OP_3(Y))
\end{align*}

\begin{align*}
\text{$\model^{O}$:} \quad \mechanism{P}(OP_1, OP_2, X, B, OP_3, Y) &= OP_1 \ (OP_2(X) \ B \ OP_3(Y)), & \mechanism{O}(P) &= P
\end{align*}
\end{center}

\end{document}